\documentclass[12pt]{article} 

\usepackage{amssymb}
\usepackage[colorlinks=true, urlcolor=blue, linkcolor=blue, citecolor=blue]{hyperref}

\newcommand{\forces}{\parallel \! \rightarrow}
\newcommand{\pforces}{\parallel_p \!\!\!\! \rightarrow}
\newcommand{\Forces}{\parallel \! \Rightarrow}
\newcommand{\can}{\mbox{ {\sf can}} } 
\newcommand{\cocan}{\mbox{ {\sf cocan} }} 
\newcommand{\coopcan}{\mbox{ {\sf coopcan} }}
\newcommand{\cocoopcan}{\mbox{ {\sf co-coopcan} }}
\newcommand{\ocan}{\mbox{ {\sf o-can} }}
\newcommand{\scan}{\mbox{ {\sf s-can} }}

\newcommand{\plans}{\mbox{ {\sf plans} }}
\newcommand{\coplans}{\mbox{ {\sf co-plans} }}

\begin{document}

\title{\vspace{-1cm}Conceptual Logical Foundations of \\
Artificial Social Intelligence}

\author{Eric Werner\thanks{
\copyright Werner 2025.  All  rights reserved. Adapted from  \cite{WernerFDAI}.    Thanking  Jon Barwise, J. Habermas, and S.C. Kleene. 
} \\
Oxford Advanced Research Foundation\\
eric.werner@oarf.org\\
\url{https://www.oarf.org}\\
\url{https://www.ericwerner.com}\\
}

\date{ } 

\maketitle

\thispagestyle{empty}

\begin{abstract}
What makes a society possible at all? How is coordination and cooperation in social
activity possible?  What is the minimal mental architecture of a social agent? 
How is the information about the state of the
world related to the agents intentions?  How are the intentions of
agents related?  What role does communication play in this
coordination process?
This essay explores the conceptual
and logical foundations of  artificial social intelligence in the context of a society of multiple agents
that communicate and cooperate to achieve some end.  An attempt is made to provide an introduction
to some of the key concepts, their formal definitions and their interrelationships.  These include the notion
of a changing social world of multiple agents.  The logic of social intelligence goes 
beyond classical logic by linking information with strategic thought. 
 A minimal architecture of social agents 
is presented.  The agents have
different dynamically changing, possible choices and abilities.  The
agents also have  uncertainty, lacking perfect information about their
physical state as well as their dynamic social state.   The social
state of an agent includes the intentional state of that agent, as
well as, that agent's representation of the intentional states of
other agents. Furthermore, it includes the evaluations agents make
of their physical and social condition.  Communication, semantic and pragmatic meaning
and their relationship to intention and information states are
investigated. The logic of agent abilities and intentions are
motivated and formalized. The entropy of group strategic states is
defined.
\end{abstract}

 
\pagebreak
\tableofcontents
\pagebreak

\section{Introduction}
Present work in artificial intelligence \cite{Attention2017, chatgpt2023, claude2024, grok2024, gemini2024, huggingface_peft_2022, chowdhery2022palm, jiang2023mistral, llama2023, llama3_2024, deepseek2024} is still limited by implicit reliance on an older paradigm of classical logic. 
While very powerful, its capacity can be greatly extended by including 
the architecture of the social mind into its remit\cite{Zhong2025, WernerTark, WernerUniView}.  Such an architecture is the bedrock of  Artificial General Social Intelligence (AGSI). To achieve this we need to link declarative reasoning about agent information states  (knowledge about the world) with strategic reasoning involving the intentional states of agents. This enables strategic rationality and action in cooperative multi-agent societies.  Integrating with this linguistic communication \cite{ WernerTark, WernerUniTheory, WernerColing} then we are at the doorstep of creating fully functional cooperative artificially intelligent social agents. Thus, for the construction of fully socially capable agents, a deeper understanding of the functional architecture of  communicative social minds may serve as a guiding roadmap. 

\subsection{Logic and Reductionism}

Logic in this century has been primarily concerned with the
foundations of mathematics, the formalization of the mathematical
concepts.   And, mathematics was the child of physics focussing on
the properties and relations of inanimate objects.  In fact, the
century started with the attempt by Russell and Whitehead to derive
all of mathematics from logic~\cite{Principia}). Later
Tarski~\cite{Tarski56} and G\"{o}del~\cite{Goedel31} showed this
reductionist program was not possible. Yet, they also restricted
their focus to the non-animate world, and, thereby, excluded the
observing subject. The mental and social realm of intentions was not
real science, it had to be eliminated. Behaviorism and reductionism
were two related philosophical movements that tried to reduce social
concepts to physical, behavioral concepts. Chomsky~\cite{Chomsky69},
and Putnam~\cite{Putnam75} and others, influenced in part by the
early developments in computer science, began to realize that
psychological states could not be reduced to behavioral properties,
but instead required functional mental states.  Thus, there was a
movement by formalists from a pure concern with the physical world
to understanding human psychology. Cognitive science was born.

\subsection{Limitations of Cognitive Science and Artificial
Intelligence}

Cognitive science tried to take the notion of information processing
from computer science and apply it directly to human mental
processes. It tried to view the human mind as an information
processing system, a symbolic computer. With rare exceptions the
work in cognitive science was not formal but relied on vague and
sometimes strained analogies to what was happening in the field of
computer science. One saw cognitive scientists writing programs
whose input was claimed to be some analogue of perception (or the
output of some previous, hypothesized mental process) and whose
output was claimed to be a stage of mental processing.

Whether or not cognitive science as a paradigm has been successful
is open to debate. The problem is that cognitive science is
fundamentally an asocial science.  The domain of cognitive science
focussed on one individual, and his or her mental information
processing. So too, the related field of artificial intelligence
was based on a similar individualistic paradigm of human mental
processes.   Artificial intelligence tried to apply this paradigm
to enable computers to simulate human reasoning. Both cognitive
science and artificial intelligence left out of consideration the
social being and the social space in which social beings interact.
They effectively ignored the social and economic sciences.  Granted
there were many exceptions (including Winograd~\cite{Winograd72},
Schank~\cite{Schank77}, Minsky~\cite{Minsky86}), but I am addressing
the paradigm, the conceptual structure underlying logic, cognitive
science, and artificial intelligence.  

This paradigm is breaking up. Information processing is just one
level or view, a rather low level view, of mental activity. It may
not even be the most appropriate way to understand this activity.
And, indeed, the information processing paradigm cannot account for
most mental and social events.

\subsection{A New Paradigm} 

The new paradigm is that of multi-agent systems science.
Multi-agent science as a paradigm studies not just the physical
world, nor agents in isolation, but the agent as part of a social
space of other agents. The information processing paradigm tended to
place the bits of information at the center of investigation,
irrespective of their meaning (much like Shannon~\cite{Shannon48}).
The multi-agent paradigm places {\em social information} and {\em
communication} at the center. Social information includes
information about values, strategies and intentions, as well as,
information about the state of the world.  This essay is an attempt
to introduce some of the concepts and the logical foundations for
the science of social information and multi-agent systems.

The importance of the new paradigm is evident when one considers
that the world is undergoing a quite radical transformation from an
information processing society where the information is acted on,
to an extended social space that includes computers as part of its
fabric.  Here the information is not individualistic and passive,
as it is in the paradigm of cognitive science. Instead, information
becomes active social information that governs and constitutively
creates the interactions between agents. Computers and robots are
becoming part of the network of social life. The image of a
single, intelligent mind in a vat, that answers questions, or a
passive machine that mimics answers to pass the Turing test, is
replaced by an active agent, a participating social being that
contributes to the social space of interactions. The information
highway coordinated by agents, and intelligent assistants are just
indicators of the beginnings of this fundamental transition of
social space and what constitutes a social being. It  opens up new,
problematic ethical issues and challenges the idea of a person as
necessarily being human.

\subsection{Comparisons to Other Formal Work}

One of the first persons to envision the new paradigm and to
understand that it was a new way of thinking was Carl
Hewitt~\cite{Hewitt77}. Hewitt realized that the interaction
between agents was a new way of programming and understanding
complex systems.  Along the lines of Hewitt's paradigm the most
significant work on agent control was probably that of Davis and
Smith~\cite{Smith80,DavisSmith83}. Others were approaching the new
paradigm from a different direction by developing a new programming
methodology they called distributed problem solving (see
Corkill~\cite{Corkill79}, Lesser~\cite{LesserCorkill81},
\cite{CorkillLesser83} and Durfee~\cite{DurfeeLesCork87,Durfee88}).
This empirically valuable work can be used as a springboard for
theoretical work in distributed artificial intelligence.  
Barwise and Perry~\cite{BarwisePerry83} helped to break up the
paradigm of standard logic and its semantics. They made the
implementation of linguistic meaning appear to be possible. The
work of Barwise provided a basis for the referential component of
multi-agent communication theory~\cite{WernerUniTheory,WernerTark}.

One cannot ignore the man whose ideas really preceded all of the
above work. In 1921 John von Neumann wrote a mathematical essay
formalizing games~\cite{vonNeumannWorks}. It's significance to
the most rigorous of the social sciences was displayed with the
publication of his book~\cite{vonNeumannMorgenstern47} linking
game theory and economics. In spite of its vision in seeing the
importance of agent interaction and in providing a formal
conceptual framework for studying such interactions, von
Neumann's work ignored communication and its relation to
cooperation. He also ignored the cognitive structure of the agent
and thus ignored the details of the social, interactive processes
of coordination, cooperation and communication. In some elegant
formal work, Rosenschein~\cite{Rosenschein86} applied some of von
Neumann's ideas to multi-agent systems. Like von Neumann, this
early work ignored communication in multi-agent cooperation, and
limited itself to the study of agent interactions as regulated by
economic rationality based on utilities. However, his more recent
work \cite{RosenZlotkin94} does address simple protocols of
interaction. Also interesting is the use of voting in establishing
agent consensus~\cite{EphratiRosen92}.

Bratman~\cite{Bratman87} gives a good philosophical account of
intentions and of reasoning about intentions.  The work is
nonformal, but insightful and well worth reading. Perhaps the
best known attempt to formalize intentions is that of Cohen and
Levesque~\cite{CohenLevesque87}.  However, there are problems
with their theory~\cite{WernerEcai90,SinghCrit92}.
Kraus~\cite{GroszKraus93} has taken a syntactic approach to
formalizing intentions. In spite of there being no semantics, I
find her work interesting because of the detail of its syntactic
formalization. Rau and Georgeff~\cite{RauGeorgeff91} formalize
the notion of intention in a somewhat similar approach to my own.
However, in their formalization, they use accessibility relations
for intentions that give little insight as to the relation of
intentions to plans or to state information. This contrasts with
their actual implementation of multi-agent planning and team
activity, which mirrors my own theory of cooperation through the
coordination of intentional states
(see~\cite{Kinny94,WernerTark}).

A number of investigators have been working on the problem of
formalizing the concept of ability. Brown~\cite{Brown88}
formalizes ability but ignores plans and strategies taking the
perspective of an agent making one choice without the
consideration of time. Singh~\cite{Singh90} tries to give
semantics of ability but fails to relate abilities to plans and
strategies, a serious shortcoming. He does give an interesting,
though convoluted, account of skills, and more permanent
abilities. Thomas~\cite{BThomas93} introduces time into the syntax, but
apparently unaware of my work, repeats some my studies, yet at the
same times suffers from not including strategies into the
semantics of ability. All of the above authors restrict their
theories of ability to the next time step. Such theories are
unworkable in multi-agent environments because they do not have a
strategic theory of ability~\cite{WernerEcai90,WernerUniView}.
Wooldridge and Fisher~\cite{WooldridgeFisher92} extend my theory
of strategic ability to include quantifiers.

Many investigators are beginning to see the importance of
including time into the syntax and semantics of modal operators
(as done originally in~\cite{Werner74} available
as~\cite{WernerMLGames,WernerTensedML,WernerPlanUnc}). Another
problem is that none of the above authors integrate state
information with strategic information. And, none of the above
authors have a formal theory of uncertainty about information and
intentions.	Yet this is crucial for understanding multi-agent
communication as an interactive process. The representation of such
uncertainties will be one of the concerns of this essay.

While there are many authors who see concepts from the social
sciences are relevant to DAI and MAS, the only authors to do formal
investigations into the logic of political interactions are the
Castelfranchi group~\cite{Castelfranchi90,Castelfranchi92} and
Werner~\cite{WernerEuroNotebooks}.  

My own theory of multi-agent systems developed out a fundamental
question: What makes a society possible at all? This came as a
natural development out of my study of communication and
cooperation from a rigorous, formal perspective. The origins of
my work began in 1970, when I received a fellowship to do research
in Germany.  I was interested in what I termed {\em open
systems}, meaning dynamic systems of interacting agents.  At
first, I was interested in the dynamics of agent choices in a
dynamic world of other agents.  My methodology was to use logic
and formal semantics. This led to a formalization of information
and strategic action that turned out to have close links with von
Neumann's work. I had developed the modal logic of
games~\cite{WernerMLGames}. Later, as discussed below, the
objective was to understand and formalize communication and
cooperation in simple social systems. This led to the development
of a general theory of multi-agent communication and social
organization~\cite{WernerTark}.  It also resulted in a
formalization of some of the Wittgensteinian language
games~\cite{WittBlueBrown,WittInv,WernerUniTheory}.

\subsection{Outline of things to come}

This essay is a somewhat personal history and perspective of this
new field. It focuses on concepts and does not attempt to give a
detailed review of all of the literature. Most of the ideas
presented here were developed by myself. This is not to say that
others have not independently discovered or rediscovered some of
the ideas presented here.

\subsubsection{Brief Overview}

We start by considering how communication may have evolved, how
meaning and understanding evolved.  The communication architecture
of an agent is then described. Next, a conceptual, critical history
of formal semantics (including Tarski, Kripke, Montague and
Situation Semantics) is given in the light of the new theory of
meaning, understanding and communication. Then, we dive into the
deep subterranean world of types of uncertainty and their
formalization.  Simultaneous actions are explored.  State
uncertainty and uncertainty about plans are rigorously
investigated. After the description of information states and the
intentional states of individual agents and groups, we go on to an
exploration of types of agent abilities, the logic of 'can' or 'is
able to'.  We also present a novel, practical theory of utilitarian
and probabilistic ability.  Next, we look at types of intention in
different social contexts.  We conclude with an	exploration of the
concept of social entropy of organizations and groups.

\subsubsection{Advice to the Reader}

While the themes may appear to be diverse, they are all actually
interconnected and necessary for an understanding of multi-agent
interaction.  One of my main goals in introducing these concepts is
for you, the reader, to gain an view of these interconnections, and,
thereby, help you to understand this fascinating field.

In terms of presentation style, I try to strike a balance between
intuitive, informal descriptions of some of the key foundational
concepts, one the one hand, and more rigorous formal definitions of
these concepts, on the other hand. Thus, intuitive, philosophical
and historical conceptual description is mixed with a rigorous
development of key concepts.  For those who find the formal terrain
difficult to traverse, please do not be intimidated and concentrate
on the ideas, which are the basis of the formal concepts anyway. You
can choose to skip or just briefly view the formal parts of the
chapter on the first reading.   Those who want a longer, deeper and
easier introduction to the field as well as more details on the
formalism, will hopefully enjoy my upcoming book on these and other
issues.

\section{The Evolution of Communication}

My fundamental assumption is that coordination and cooperation is
only possible by way of the exchange of information.  Isolated
agents that are truly autonomous with no interactions are assumed to
have no coordinated behavior.  There may be relationships between
their behavior because they may have a similar structure or
experience. However, similar structure is a result of the phylogeny
of the organism in its interaction over generations with the
environment.  Similar experience is a result of the ontogeny of the
organism resulting from the interactions of the organism with its
environment over its lifetime. In both cases the environment can be
viewed as another agent, an agent common to both our original
agents.  Their common structure is then a result of an interaction
with the same environment.

\subsection{The Solipsistic and Alien Agents}

A {\em solipsistic agent} is a single agent with no interaction
with other agents or with the environment because it accepts no
input. Such an agent may be difficult to imagine, but we include it
for the sake of completeness in our logical space of possible agents
as classified by their interactions. Such an agent may be a robot
gone out of control having lost its senses and acting only on its
internal imagined representations.  Note, such an agent is still
phylogenetically speaking couched in its environment. Even its
ontogenetic structure may be intact.  An {\em alien agent} is one
whose phylogeny and ontogeny is not from the multi-agent world in
question and with a concomitant lack of commonality of ways of
being with and understanding the world.  Thus, its degree of
alienation is relative to a multi-agent system which in turn is part
of a more general ecology.

\subsection{Reactive Agents and Reactive Conversations}

A {\em purely reactive agent} is one whose strategic state is
purely reactive to its environment.  If there are other agents in
the environment it treats them on a par with other objects and not
as harbingers of meaning.  As such agents gain the faculty of
memory, their reactions can become more and more sophisticated.
And, coordination may even be possible if the reactive strategies of
the agents have coevolved so as to achieve some social end.  The
coordination in this case was achieved with Mother Nature being the
common agent of interaction.  Mother Nature weeded out all the
reactive strategies that did not fit together.  Consciously
programming such agents to have coordination 'emerge' can be very
difficult. Since the agents change the properties of the world in
some way, these properties if unique enough may be a stimulus for
another agent including the agent who generated the stimulus.  In
this way a {\em reactive conversation} may emerge.  One particular
and popular beginning programming project is to have the agents
interact by leaving a scent~\cite{HussmannWerner90} or dropping
pebbles. In these cases the environment can be viewed as being the
program that the agents react to.  The agents program each other
and themselves by changing their common environment.

\subsection{User Guided Evolution of Reactive Coordination}

A different approach to programming agents can be taken by
extending the work of  Karl Sims~\cite{Sims93} as motivated by
Daniel Hillis~\cite{Hillis90} and Holland~\cite{Holland75}.  Transferring
and slightly generalizing these thoughts on graphics to agents,
let the agents strategies evolve over time and let the user play
Mother Nature in selecting those reactive strategies that suit
him.  Here the coordination that evolves is the result of the user
as common agent. The user's tastes influence the phylogenically
induced strategic control structure of the agents.  Reactive
interactions that are coordinated in space time may evolve quite
naturally instead of being coerced  by the more conscious and
painful efforts of a programmer who directly and explicitly tries
to design and program coordinated single and multi-agent
interaction strategies.

Simple reactive agents are extremely limited in the kinds of
coordination that they can achieve.  Real coordination and
cooperation only become possible if we have communication.

\subsection{Where does meaning in communication come from?}

Historically, attempts at a theory of meaning have been dominated by
referential theories of meaning.  These are theories that view
meaning as a relationship between the language as syntax and the
world as a set of objects (as well as properties and relations, in
more sophisticated later versions).  The meaning of a language token
is what it refers to.  For example, "Mouse" refers to a mouse.  "On
top of" refers to a relation. The point is that an object or event,
e.g., a sound or a stone, or a picture, represents something else.
And the information content of this something else is usually much
greater than and/or different from its symbolic representative
object.  This is the core of referential communication. It allows
the transfer of information about an object, situation or event
without transferring all the information of the actual object or
its original perception.

\subsection{What makes communication possible?}

How is such a transfer possible?  How can we transfer information
in a conduit that is clearly incapable of actually representing the
information it is to transfer?  The answer lies in the nature of
understanding.  To understand a symbolic communicative object or
event (e.g., a symbol, sound or gesture) means that the subject
interprets the symbol.  The subject has a capacity to relate the
symbol to previously stored information.  The symbol's meaning lies
in its association with a representation in the subject and how the
subject transforms his representation of the world in response to
the symbol's representation.  This is what I have called the
pragmatic meaning of a symbolic object or event.

\subsection{Pragmatic Meaning}

Meaning, on this view, is not a passive representation that
pictures the original object.  Instead, it is associated with an
operator that transforms the mental state of the subject.  The
subject brings more or different information to the meaning and
interpretation of the symbol than can ever be informationally
possible in the symbol, as conduit, itself.  If we take the
subject's information state $I$ to be his representation of the
state of the world, then $I$ gives partial information about the
world that is gotten by direct perception and by communication.  A
symbol $\alpha$ communicated to the agent transforms the agent's
information state $I$ to a new state $J$ where $J = \alpha I$.
Here, the symbol $\alpha$ acts as an operator on the receiver's
information state.

The point of the above philosophical discussion is that if $\alpha$
is just a syntactic symbol, or noise, or gesture, it could have any
arbitrary meaning. For to have its effect as an operator the
symbolic object or event $\alpha$ must be given a pragmatic
interpretation by the subject.  If our subject is Dany, then let
$Prag_{\sf Dany}(\alpha)$ be Dany's subjective interpretation of
$\alpha$.  If Dany accepts the message $\alpha$, say that Sophie is
in the kitchen,  then the message $\alpha$ transforms her
representation to one where Sophie is in the kitchen. Dany's
information state becomes $J_{\sf Dany} = {\mathit Prag}(\alpha)(
I_{\sf Dany} )$ which says that the Dany's new information state
$J_{\sf Dany}$ results from her previous information state $I_{\sf Dany}$ 
as transformed by the operator ${\mathit Prag}(\alpha)$ which is
the pragmatic interpretation of the message $\alpha$.  It is
actually a bit more complex, because the pragmatic interpretation
function {\em Prag} is relativized to Dany, ${\mathit Prag}_{\mathit Dany}$
since Dany's interpretation of messages like $\alpha$ may differ
from that of other agents.

\subsection{Referential Agents}

Given that reactive agents have the ability to partially represent
the world, at least one capacity that distinguishes communicating
agents from purely reactive agents is the capacity to interpret
messages pragmatically in the above sense.  Such a capacity will
give the agents a capacity to engage in referential pragmatic
interpretations of symbols. A referential pragmatic interpretation
changes the information state $I_A$ of the agent $A$. Let us call
such agents {\em referential agents}.

What do referential agents gain from such pragmatic capacity?  We
will see below that information about state increases the
capabilities of an agent as well as the capabilities of groups.
Furthermore, in so far as vision is an interpreting process of
visual information, vision itself can be taken to be a pragmatic
operator that interprets visual information and transforms the
information state of the seeing subject. Vision is in this sense
like a language that transforms the information state $I_A$ of
the observer-agent $A$. Like a language we learn to see more with
experience.  With learning, the pragmatic visual operator changes.
In fact, the visual pragmatic interpretive capacity, as well as,
the cognitive capacities this presupposes, may have been necessary
for the evolution of the linguistic pragmatic interpretive
capacity.    

\subsection{Intentional Agents}

If all that agents could do was to talk about the state of the
world, in the sense that their pragmatic operators were restricted
to referential interpretations, then agents might be able to
describe and to react to this information.  For example, an agent
may report a bear is coming toward the cave and the agents might
react to this new information state by running away, given they have
evolved such a strategy.  (The last agent who did not have such a
reactive strategy to bears approaching their caves died of bear
wounds.  Perhaps this was the fate of the poor Neanderthal who is
reputed to have had a brain larger than our own.)  However, if the
agents wanted to kill a mammoth they may have had to coordinate
their activities.  And, reactive strategies to communication of
state information will in general not be sufficient to elicit the
complex coordination required in hunting activities.

Coordination requires something not found in a referential account
of meaning.  It requires the coordinated formation of strategic
information states or intentions.  By a {\em strategic information
state} or {\em intentional state} I mean a mental state of the agent
that represents and constitutes her control state. It is the state
that guides and controls her actions in the world.

\subsection{Evolution of Communication about Intentions}

In itself, without communicative interactions, a control state may
be reactive, as well as, being tactical and strategic.  It may have
sophisticated planning capabilities associated with it.  This is
much like animals who have both reactive, tactical, strategic and
planning capabilities.  A key event in the evolution of meaning was
when symbols began not just to have referential meanings, but to
have strategic pragmatic meaning.  A symbol (e.g., sound, object,
picture, text) began to be associated with an action as well as
with whole strategies of action.  These symbols required a
different sort of pragmatic interpretation.  Instead of changing the
information state, the operator associated with the symbol
transformed the strategic control state of the agent.  It
transformed her intentions.  And, by transforming her intentions it
allowed the agent's actions to be directly influenced, controlled
and coordinated.  This fundamental development was crucial to all
further human and animal organization.  And, it is the basis for
complex coordination and cooperation in multi-agent systems
generally.

To systematically distinguish a message that has a primarily
strategic meaning let us use the exclamation mark '!' to indicate a
request or command. Thus, if $\beta !$ is a strategic symbol, the
pragmatic interpretation of $\beta !$ is ${\mathit Prag}_A( \beta ! )$
where $A$ is the receiving, interpreting agent.  We assume that the
agent has a capacity to represent its intentions.  Let us call this
representation the agent $A$'s {\em intentional state} and
symbolically denote it as  $S_A$.

\subsection{The Architecture of Social Agents}

Social agents have not only intentions that guide their actions
but also the ability to take on various roles in social settings.
These may range from very particular roles in action situations to
complex responsibilities and permissions in an organization. As
described in my work on communication and social
structure~\cite{WernerTark,WernerUniTheory} such roles can involve
both state information $I$, strategic information $S$ and
evaluative information $V$. Together these basic types of
information constitute the agent's {\em representational state} $R
= ( I, S, V )$. The adoption of a role then forms and constrains
the agent's overall representational state.  In order to be able to
adopt a role the agent must have the abilities required by the role
(see the section on abilities below).

\subsection{Communication Theory and the Force of Speech Acts}

In the theory of communication that I
developed~\cite{WernerTark,WernerUniTheory}, I distinguished
between the syntax, the semantics and the pragmatics of a language.
The {\em syntax} consists of a set of rules that generate the
grammatical sentences of the language. The {\em semantics} relates
language to situations in the world or representations of
situations in the world.  The {\em pragmatics} defines how language
transforms the mental state of the communicators.  The speaker's
pragmatic competence is given by a compositional semantics and
pragmatics that gives a semantic and pragmatic meaning to every
sentence of the language generated by the syntax.	This is
different from the categorization made by
Morris~\cite{Morris46,Morris64,Morris72}.

For those familiar with speech act theory as developed by
Austin~\cite{AustinSense,AustinHow} and Searle~\cite{Searle71},
the pragmatics of a language, in my sense of the term, is not a
theory of the perlocutionary effects as those are more like side
effects of an utterance.  The pragmatics is, instead, a theory of
how language manipulates the representational state of the agent.
For those not familiar with speech act theory, a speech act
consists of a force plus a propositional content.  The force
indicates whether we are dealing with a request or assertion, for
example. The propositional content is the referential component of
the speech act that links it to the world.	For example, 'Open the
door!' and 'The door is open.' are said to have the same
propositional content but different force.

The theory of representations and pragmatic meaning outlined above
allows us to give a novel and noncircular definition of the force
of a speech act~\cite{WernerColing}.  The {\em force} of a speech
act is the focus of the pragmatic transformation.  This focus can be
directed at one of the following three fundamental types:  The
information state, the intentional state or the evaluative state.
Thus, for example, assertives such as, 'He is in the kitchen', focus
on the information state while commands  and requests like , 'Please
give me some more cable', focus on the strategic, intentional
state. Evaluatives or emotives like, 'I like that', focus on the
evaluative state.  Thus, the force of a speech act indicates the
dominant part of the representational state of the agent that is to
be changed by that speech act's pragmatic meaning, given that the
message is accepted by the receiver. This theory of force has
several advantages over the previous theory of Vandervecken and
Searle as their definition is essentially circular, syntactic with
no semantics, and does not relate force to mental states in a
systematic way (see~\cite{WernerColing} for details). It should be
obvious that this notion of speech act force has nothing to do with
the notion of agent power.

\section{Meaning in Multi-Agent \mbox{Systems}}

\subsection{The Meta-Observer and the Object System}

The designer of a multi-agent system $\Gamma$ as observer and agent
can and usually does take a meta perspective with regard to that
system.  By {\em meta-perspective} I mean that the system is viewed
and acted on from the outside, from a perspective and action space
external to the system.  Let us call the agent that is observing the
system from the meta-perspective the {\em meta-observer} and the
observed system the {\em object system}.  The agents in the object
multi-agent system will be called {\em object agents}. The {\em
object world} is the world in which the agents act.  This world
contains the object agents as well. The {\em state} of a multi-agent
object system minimally consists of the object space-time world
state (physical environment), the state of the agents as objects in
that world, as well as the internal (informational, intentional and
evaluational) state  of the agents.

\subsection{Object and Meta-Language}

Our notion of meta-perspective was motivated by the logician
Alfred Tarski.  Tarski~\cite{Tarski56} made a distinction between
the {\em object language} that the logician is investigating,
e.g., the formulas of first order logic, and the {\em
meta-language} that the logician uses to talk and prove things
about the object language. The typical meta-language was a natural
language such as English supplemented with a mathematical language
such as set theory.  A proof in the object language was different
than a proof in the meta-language.  The meta-language is typically
more powerful than the object language.  This allows the object
language to be translated into the meta-language. Hence, to each
term $\bar{\alpha}$ in the object language there corresponds a
term $\alpha$ in the meta-language.

\subsection{Multi-Agent Object and Meta-Languages}

We propose making an analogous distinction for the study of
multi-agent systems.  Hence, we distinguish the {\em agent object
language} ({\sf AOL}) that the agent's use to communicate, from the
designer's {\em agent meta-language} ({\sf AML}), that the system
designer uses to talk about the agents and their communication.
Distinguishing the {\sf AOL}  from the meta-language has the
important consequence that it keeps the communication interface
between the agents precise.  Agent linguistic communication is
restricted to the agent object language {\sf AOL}.  When this is not
done one can observe that the programmer designing and implementing
the system has a difficult time maintaining the boundaries between
his program (the meta-language) and the interaction language
between the agents.  Indeed, most often the programmer is totally
unaware that he is mixing the two.  The result is a possibly poor,
messy design that is not modular and also difficult to extend and
maintain.

\subsection{Object Semantics and Object Pragmatics}

As we saw, {\em semantics} relates language to the world, {\em
pragmatics} relates language to how it transforms the mental state
of the agent. The pragmatics is realized through {\em pragmatics
operators} that act on the information state, intentional state and
evaluative state of the agent.  A pragmatic interpretation of a
language associates operators with sentences of the language.  For
each sentence $\alpha$ of the language the associated operator {\em
Prag}$(\alpha)$ indicates how the intentional state, the information
state, and the evaluative state of the agent is to be updated or
transformed (see~\cite{WernerUniTheory} for more details).

For Tarski, semantics was part of the meta-language.  He took the
role of the meta-observer who provides a theory of meaning for the
given object language by giving it a formal semantics where the
terminology used to describe the semantics was part of the
meta-language.  He did not consider agents or their communication.
He was concerned with the semantics of logical formulas.  We
however, are interested in understanding the process of
communication between agents.  In communication it is the {\em
agent} who interprets and gives meaning to the agent object language
{\sf AOL}.   So, for us, the semantics and pragmatics become part
of the agent, and thus, they become part of the agent as objects of
meta-observation.

Therefore, from our meta-perspective (you the reader and I), we
distinguish the meta-language and its {\em meta-semantics} used by
an observer of a multi-agent system $\Gamma$ from the agent object
language {\sf AOL} with its associated {\em object-semantics} {\it
OS}$_B$  and {\em object pragmatics} ${\it OP}_A$ which is used by
the object agents within that system $\Gamma$.  Thus, if we want to
be very precise, consider a meta-agent $A$ observing a multi-agent
system $\Gamma$ containing an object agent $B$.  We can distinguish
between the agent $A$'s meta-language {\sf AML}$_A$ and its
associated meta-semantics {\it MS}$_A$, from the agent object
language {\sf AOL}$_{\Gamma}$ and the associated object semantics
{\it OS}$_B$ and object pragmatics {\it OP}$_B$ for {\sf
AOL}$_{\Gamma}$.

For example, the designer, and possibly the user, can communicate
with the agents at their level if he communicates with those agents
using the agent object language.  The designer must interpret the
agent object language, by way of the object semantics, in terms of a
meta-linguistic description and its corollary meta-interpretation.

The situation is a bit complicated because the object semantics
when studied by the meta-observer is described in the
meta-language of the meta-observer.  And, this meta-language is
meaningful to the meta-observer because he has a meta-semantics
and pragmatics to interpret the meta-language.  These are the
considerations that arise when we (you the reader and I) are the
meta-meta-observers observing the combined system of meta-observer
(now as object agent relative to us) observing an object agent
community.  As an aside, it is interesting to note that this is
just the situation in Everett's many-worlds interpretation of
quantum mechanics where Everett assumes the role of
meta-meta-observer and adds the meta- observer (which von Neumann
kept separate from the observed system) to be part of the system
under observation (see~\cite{vonNeumann55,Everett57,Wheeler57}).

The notion of {\em object semantics/pragmatics} is needed because
the capacity of the agent to understand the agent object language
{\sf AOL} is given locally by the agent's object structure. This
concept is important since the object semantics and pragmatics is
local to the agent.  The agent object language together with its
object semantics and pragmatics encapsulate the linguistic
communicative properties of the agent.  And, thereby, allow a well
defined communication interface to the agent.  They make the agent
part of a linguistic community.

If we imagine a group of robots building a space station, their
agent object language might be a simplified natural language of
directives, requests, commands, informatives, and simple
evaluatives. The robots would have an object semantics and
pragmatics to interpret this object language. Furthermore, they
would be able to execute the new intentional states that result
from the communicative interchange.  That is, there are two levels
of interpretation happening.  One level interprets the object
language with the object pragmatics by acting on the agent's
intentions (plan states).  The level below is an {\em object plan
interpreter}, which may be local to the agent, that executes the
active intentional state.

\subsection{The Relativity of Semantics and Pragmatics}

Traditionally, formal theories of meaning, such as model theory
or possible world semantics, have failed to distinguish the
object semantics from the meta-semantics.  Even Tarski, who first
made the distinction of the object and meta-language failed to
recognize the distinction of object and meta-semantics.  This
failure resulted in the implicit projection, of part or all of the
meta-semantics, onto all agents as the universal object semantics.
That meant that within these theoretical frameworks one could not
give an account of inter- and intra-community linguistic variance
in meaning and understanding. Both Tarski semantics, as well as
possible world semantics, are not defined with regard to agents.
The point of view, i.e., the meta-perspective, is external and
global.  It is that of a meta-observer having a clear and correct
understanding of the semantics.  As a consequence, they could not
explain misunderstanding.  Once the distinction between object
and meta-semantics is made, however, it becomes clear that there
may be variance and divergence in the object semantics/pragmatics
within and between communities of agents.  Furthermore, we are now
in a position to give a precise account of inter- and
intra-community variance of language meaning and understanding.

In a multi-agent system, each agent has his own realization of the
object semantics and object pragmatics of the agent object language
{\sf AOL}.  It is, therefore, possible that each agent has
different object semantics and pragmatics from every other agent.
This means that the pragmatic operators {\em Prag}$_{A}(\alpha)$
are relativized to the agent $A$ that is interpreting the sentence
$\alpha$ of the agent object language {\sf AOL}.  It can happen
that for two agents $A$ and $B$, it is possible that for the same
sentence $\alpha$ in their common agent object language the
pragmatic interpretations differ: 
\begin{eqnarray} {\mathit Prag}_{A}(\alpha) \neq {\mathit Prag}_{B}(\alpha)
\end{eqnarray}  
Thus, each agent has his own local
interpretation of any given message $\alpha$.

\subsection{Misunderstanding and Learning a Language}

It is a different question as to how the commonality and
divergence of meaning arises. The relativization of the object
semantics and pragmatics to agents makes misunderstanding
possible.  Now, as we would like our agents to understand each
other, one might also want an agent to be gradually integrated
into a society of agents by learning their norms, roles, and their
language ( see~\cite{WernerOntogeny}). But, that implies that the
agent starts out in a state where he only has a partial
understanding of that language. Hence, the agent's starting object
semantics and object pragmatics may be different from that of the
community.

A theory of language learning needs to give an account of how the
object pragmatics of the learning agent, e.g., a child, changes to
be in line with the object pragmatics of the community.  Such a
theory requires a relativization of the semantics and pragmatics to
the agent. This relativity of semantics and pragmatics to the agent
is just a natural extension of the relativity of knowledge of the
agent. For, each agent $A$ has his own information state $I_A$,
intentional state $S_A$, and evaluative state $V_A$.  And, these
states are dynamic, changing with time as the agent interacts with
other agents in the world.  The relativization of semantics and
pragmatics to the agent allows the object semantics and pragmatics
to be dynamic as well. 

The communicative interaction of agent and social context results
in a gradual acquisition of the capacity to understand the
language; the agent learns the semantics and pragmatics of the
language.   Note, Chompsky~\cite{Chomsky69} really only addressed
the problem of how speakers learn the syntax of a language.  Here
we are confronting the problem of how speakers learn to understand
a language.  Winograd in his early work on a robot controlled by
language clearly saw the difference~\cite{Winograd72}  as did
Schank~\cite{Schank77}.

\section{A General Communication Architecture {\sf ICE}}

Based on the above conceptual developments, let us outline the
general architecture of agents with a communicative competence.  We
call it the {\sf ICE} architecture which is short for ${\sf
I^{2}C^{2}E^{2}}$ or  {\sf I}nformation, {\sf I}ntention, {\sf
C}ommunication, {\sf C}ooperation, {\sf E}valuation, and {\sf
E}mpowerment architecture for dynamic interacting agents.

\subsection{Agent Cognitive Architecture}

The {\sf ICE} architecture for the mental  or {\em representational 
state} of an agent $A$ consists of tuple $R_A= (I_A, S_A, V_A)$ 
where:

\begin{enumerate}

\item $I_A$ is the agent's {\em information state}.  We allow
nesting. $I_{A}^B$ is $A$'s state information about $B$'s
information. $I_A^A$ is $A$'s information about its own information
state.  The information state gives the agent's information about
its world at a given time.

\item $S_A$ is the agent's {\em strategic or intentional state}.
Again we allow nesting as with information states.  It consists of a
partial strategy or, equivalently, a set of strategies that guide
the agent's choices of action.

\item $V_A$ is the {\em evaluation state} of the agent.  It can
consist of a utility function over objects, situations and states.
It may be discreet, indicating wants or desires.

\end{enumerate}

The representational state $R$ and its components $I$, $S$, and $V$
are all dynamic changing in time and modified by the dynamic
interactions with the environment and with other agents.

\subsection{The Space of Possible Representations of an Agent}

The agents can form concepts, plans, and evaluations
dynamically while they are in some goal directed process.  Part of
the competence of the agent's real time planning and strategic
capabilities comes from a repertoire of plans and concepts.  This
repertoire is combined dynamically to form the given actual
intentional state of the agent.
  
\begin{enumerate}

\item ${\it PC}(\Omega)_A$ is the {\em absolute plan competence
space} of the agent.  It is the set of possible partial (including
complete) plans and strategies of the agent given the action space
$\Omega$.  It sets the limits of what is logically possible for the
agent given his action space.  All competencies that follow are
within this space.

\item ${\it PR}_A^t$ is the {\em plan repertoire} also called a
{\em plan library} of the agent $A$ at time $t$.  These are the
basic skill fragments, tactics or strategies the agent has
inherited or learned and are available dynamically to construct a
much larger set of real time available dynamic plans. Note, this
repertoire is itself dynamic since the agent can learn a new skill
or strategy while attempting some goal directed process.

\item ${\it PC(PR)}_A^t$ is the {\em plan competence} of the agent
given his plan repertoire at some time $t$.  It consists of the
space of possible plans as limited by $A$'s repertoire.  It allows
unbounded computation time, but no new skill development.

\item $ \Delta(R_A)_A^t$ is the set of {\em dynamic strategies}
available to the agent in real time relative to his plan repertoire
${\it PR}_A^t$ at time $t$.  Here 'real time' depends on the time of
activity as an agent can adjust his plan state dynamically in a
longer goal directed process.   This dynamic competence will in
general depend on what the agent knows and so also depends of his
representational state $R_A$.

\end{enumerate}

We have introduced a dynamic plan space $\Delta_A$ because a plan
library does not indicate the real time plan competence of the
agent.  For, the agent may not, and need not, have a full
representation of the abstract, partial plan he is executing.  He
will usually fill in the details as he performs the plan depending
on the local circumstances, surprises, failures and opportunities.

\subsection{Communicative Competence} 

Given an agent architecture with  $R_A= (I_A, S_A, V_A)$ and a  
dynamic repertoire 
$\Delta(R_A )$  we can define 
the communicative competence of the agent:

\begin{enumerate}

\item {\bf Syntax}  The agent has a language $L$ the {\em agent
object language} {\sf AOL} with a syntax to generate the sentences
of the language.  This syntax can range from being very basic
consisting of a very few symbols to being an advanced natural
language like English.

\item {\bf Semantics} {\em OS}$_A$  The agent has an {\em object
semantics} {\em Semantics}$_A^L$ for the language $L$.  The
semantics provides a situational meaning to the basic expressions
and sentences of the language relating the language to the agent's
world.

\item {\bf Pragmatics} {\em OP}$_A$  The agent has an {\em object
pragmatics} {\it Prag}$_A^L$  for the language $L$. The agent's
object pragmatics associates with each sentence $\alpha$ of the
language $L$ an operator ${\it Prag}_A(\alpha)$ on the
representational capacity of the agent.  It takes a representational
state $R_A= (I_A, S_A, V_A)$ and transforms it to a new
representational state: 
\begin{eqnarray} 
{\it Prag}_A(\alpha)(R_A) = R_{A}'  =
({\it Prag}_A(\alpha)(I_A), {\it Prag}_A(\alpha)(S_A), {\it
Prag}_A(\alpha)(V_A))
\end{eqnarray} 

\item {\bf Linguistic Strategies}  The agent has a {\em dialogue 
competence} that allows him to interact in conversations.  These 
consist of linguistic protocols and strategies. These may be
dynamically formed from a repertoire.  

\end{enumerate}

If we let $\alpha$ be shorthand for the sentence and the operator, 
this states $\alpha(R) = \alpha R = R'$.  We can then easily chain
communication operators $\alpha_1\alpha_2\alpha_3 R$ which
represents a series of pragmatic messages starting with $\alpha_3$
then $\alpha_2$ and ending with $\alpha_1$ operating on $R$.  The
{\em force} of the message (speech act) is that subrepresentation
of $R$ that is the primary focus of the operator.  For example, a
command or request will have as its primary focus the strategic
intentional state $S_A$ with a secondary focus on the information
state $I_A$ indicating that a message has arrived, etc.  So, the
force of a command or request is strategic.  The focus of assertions
is on the information state $I_A$ and its force is informative.

\subsection{Cooperation or Social Architecture}  
Beyond a basic communicative competence to pragmatically 
understand a language the agent will have a social competence to 
interact with the language.  It is usually not recognized by linguists 
that this is a fundamental additional linguistic and social skill of an 
agent.  Additionally, the capacity to interact socially in organizations 
with roles and networks of such are additional social competencies 
that agents may or may not have.   

\begin{enumerate}

\item {\bf Social Roles}  The social agent has the capacity to
participate in, take on, and recognize social roles.  Roles include
permissions and responsibilities.  Roles may also have associated
linguistic strategies of interaction.

\item {\bf Organizational Competence}  The agent has the competence 
to functionally represent and participate in 
organizations.  Organizations consist of complexes of roles and 
suborganizations.  

\item  {\bf Networks of Organizations}  
The agent has the competence to represent and participate in networks 
of organizations.  The networks are linked by special network roles.  
These may involve special linguistic protocols.  

\end{enumerate}

\section{Motivation and Development of the {\sf ICE} architecture}

\subsection{Conceptual History of the {\sf  ICE} Architecture} 

The {\sf  ICE} architecture was developed independently in the mid
to late seventies as a basis for understanding communication and
cooperation among multiple agents.  The theory behind {\sf  ICE}
was developed in three major phases.

\subsubsection{Operators on Information States}

First, in my work on the temporal modal logic of games, I had
developed a formal dynamic semantics that included information
states as part of the semantics.  I realized that one should be
able to use the formalization of information within semantics to
develop a theory of how information is communicated by language.
More particularly, I became interested in understanding how the
information state of agents changes as they communicate. I
developed a theory of how communication changes information states.
This is what I called a pragmatic theory of the meaning of
informative messages.  The subject or agent had a mental state, his
information state and this state were transformed by the
communicative messages he received and interpreted.

\subsubsection{Relations to Observables in Quantum Mechanics}

This led to a view that meaning is like an operator on the
information state of the subject changing the information state
by operating on it to form a new information state.
Independently, I had been doing reading in quantum mechanics, and
in particular, von Neumann's mathematical foundations of quantum
mechanics and the theory of how observations act as operators on
the state (as a vector in a Hilbert space) of the wave function
\cite[von Neumann]{vonNeumann55}. By viewing the observer's state
itself as a wave function, we can view the observation as changing
the state of the observer and the observed. Physicists had
studied interesting properties of such operators and I realized
that they had analogues in normal communication. Furthermore, the
gaining of information about a phase space in statistical
mechanics, with its reduction of entropy, as the reduction of the
phase space (see for example Khinchin~\cite{Khinchin49} and
Boltzmann~\cite{BoltzmannWorks}), was similar to the way
communication operators reduce the information set defining the
information state of the agent. These similarities with other
seemingly dissimilar theories suggested that there was some
fundamental, universal aspect of information that had been
captured by the theory of information states and their operators.

\subsubsection{The Influence of Sociology and Intentional States}

This representational operator model of communication was 
extended, when, stimulated by the writings of the sociologist
J\"{u}rgen Habermas~\cite{Habermas81}, it became obvious to me that
the most important aspects of human communication, namely
communication in social cooperative activity, could not be
explained by communication about state information alone.  Some
other representational structure was needed.  Something that
controlled the agent's actions:  Control information.  So, I
developed the theory of intentional states with the explicit aim
that such states must be dynamically transformable (operated on)
by communicative interactions.  The idea was that communication
between agents dynamically form the intentions of agents setting
up their intentional states so that coordinated activity and
cooperation were possible.  This I saw as the foundation for a
theory of society.  I viewed the intentional state as that which
controls the agents activity on the basis of the evaluations and
state information the agent has about the world.

\subsubsection{Language Games}

Habermas also reminded me of my student readings of
Wittgenstein~\cite{WittInv,WittBlueBrown}. Wittgenstein had actually
made a similar discovery, that state information cannot adequately
account for all the functions of language.  In his terms the meaning
of language lies in its use. His examples of various language games
show that the meaning of the expressions used in the games, e.g.,
'Brick!' when uttered by a mason to his apprentice to mean 'Bring me
a brick!' could not be reduced to an assertion that was true or
false. They are good examples of self-contained mini social
situations of agents engaged in communication and cooperative
activity.  It is for this reason that I formalized some of his
language games, to test my theory of communication.  For one of
the simpler formalizations see~\cite{WernerUniTheory}.

\subsection{Motivations for BDI}

How does the {\sf ICE} architecture differ from the BDI (for
beliefs, desires, intentions)   architecture?  The BDI
architecture was based on the work of Bratman~\cite{Bratman87}.
It was  developed after Bratman had independently developed a
theory of  intentions. His intuitions and discoveries
about  intentions were strikingly similar to mine.  However, our
original  motivations were different. He was more interested in
capturing the  properties of human intentions and their function
in human reasoning  and decision making, whereas I was interested
in the function of intentions in communication and social
coordination.  My ultimate motivation was an abstract theory of
what makes social cooperation possible.  And, I viewed
communication as playing a central role in social cooperation.  It
should also be noted that the BDI architecture  is not a formal
theory but rather a verbal description that beliefs, desires and
intentions ought to be part of the mental states of the agent.  It
says nothing of how to precisely integrate these notions into  a
coherent theory.

\subsection{Requirements of a Theory of Animal and Robotic 
Communication}

Since I was trying to develop a general theory, I was not only
interested in human communication and cooperation but in animal and
robotic communication and cooperation  as well.  For, animals
exhibit quite  sophisticated social coordination and signaling.  A
theory of  communication and coordination should be able to explain
such  phenomenon as well.  Since they are probably simpler
phenomenon  than human communication and social coordination, I saw
them as an  interesting area to test a theory of communication and
cooperation.

If one wanted to build robots with various degrees of abilities,
communicative and social competence, then a theory of
representational states (such as the {\sf  ICE} architecture),
communication, and cooperation should be abstract enough to permit
the design of very simple communicating and cooperating robots.
Such robots may not reason symbolically and may lack sophisticated
intentions and beliefs, but they still need strategic control states
and information states that are manipulatable by some form of
primitive communication.  More generally reactive agents might have
very simple, nonsymbolic, nonanthropomorphic strategic control
states.  Yet these might still respond to a primitive, or even
complex, pragmatically interpreted language.

\subsection{Problems with Belief and Desire}

This is one reason I stayed away from the notion of belief.  The
logic and semantics of belief and knowledge has been controversial
since its inception.  Information states are more abstract and make
use of solid theory about the nature of information, control,
entropy and game theory.  Indeed, the way the 'Belief' predicate is
actually used by adherents to the BDI architecture, it is little
different than a symbolic way of indicating what state information
the agent has.  Similarly, I preferred the notion of evaluations
because, again I did not want to make anthropomorphic projections on
the evaluations of utility that animals, humans or robots may be
making.

\subsection{Universality:  BDI as a special case of {\sf  ICE}}

In a further demand for universality, my notion of intentional
state as a set of possible strategies that control an agent,
allows a formalization of control information that has nice
properties that make it fit well with a general theory of
information and communication. Also, it allows an integration of
state information, strategic-intentional information and
evaluative-utility information.  This integration allows the
explanation and formalization of interesting social phenomena.  It
is necessary to explain the relation of state information,
abilities of an agent and the intentions of the agent
\cite{WernerUniView,WernerEcai90}. Furthermore, the abstract {\sf ICE} architecture
allows the inclusion of the BDI architecture as a special case.
Namely, beliefs correspond to and are one way to define
information states, desires are particular sorts of evaluations,
and the informal notion of intention corresponds to the more
rigorous notion of plan state and intentional state. Finally, the
{\sf ICE} architecture allows but does not require a commitment
to a symbolic representation of information (including beliefs),
intentions, or evaluations (including desires). A connectionist
or other theory of mental representational states is allowed by
the {\sf  ICE} architecture.

\section{A Brief Conceptual History of Formal Semantics} 
 
We now look at the work in formal semantics that has had an 
influence on the formal study of multi-agent systems and formal 
communication theory.   

\subsection{Tarski Semantics}
 
First, there is the tradition of modern mathematical logic which
began with Boole and continued in this century with its emphasis on
formal axiomatic systems (Frege, Hilbert, Russell and
Whitehead~\cite{Principia}).  The work of Tarski is significant
because it was the first formalization of the semantics of first
order logic with quantifiers over individuals.
Tarski~\cite{Tarski56} viewed {\em semantics} as establishing a
relationship between the language of logic which he called the {\em
object language} and the world which he called a {\em model}.
Tarski distinguished the object language from the meta-language.
Recall, for Tarski, the {object language} is the language the
logician is studying.  The {\em meta-language} is the language the
logician uses to investigate the object  language. The
meta-language contains the object language but it also consists of
ordinary English and any other mathematical language that  the
logician chooses to use.  Thus, the semantics of the object
language is  described in a semi-formal meta-language. The object
language has  explicit and rigorous rules of syntax which specify
precisely what  sentences and expressions are part of the object
language and which  are not.  In contrast, the meta-language need
have no precise syntax since it is usually an informal mixture of
natural language and formal  mathematical language.
 
To be termed a {\em formal semantics}, there must be a precise
mathematical description of the world, called the {\em model},
consisting of objects, their properties and relationships.  There
must also be a precise mapping between the object language's terms
and predicates, on the one hand, and the objects, properties and
relations in the model, on the other hand. And, there must be a
precise  definition of the conditions under which any given sentence
of the  object language is true or false. The latter are called the
{\em truth conditions} (also satisfiability, validity conditions) of
the semantics.   More formally, a Tarski model for a language $L$
consists of $< O,  R^n, \Phi >$ where $O$ is a set of objects, $R^n$
is a set of properties and relations on $O$, and $\Phi$ is an
evaluation function that assigns  truth and falsity to sentences of
the object language $L$.

With Tarski's work, one starts to have a feeling or intuition that
it is the beginnings of a formal theory of meaning for a
language. In fact,  the work of Tarski has had a tremendous
influence on formal logic,  linguistics and computer science.  Its
dominant influence in computer  science and linguistics has been
by way of the semantics for modal  logic (which investigates
operators like "It is possible that", "It is  necessary that", "It
will be the case that", "A Believes that", etc.)   The syntax and
some of the axiomatic systems of modal logic were  first
investigated by Lewis~\cite{LewisLangford}. But it took
some 30 years before semantics of modal logics developed.
Hintikka, Prior, and Kripke made some ground breaking investigations
into the semantics of modal logic
(see~\cite{Hintikka62,Prior67,Kripke63}). Kripke's work is perhaps
the most abstract and generally applicable.

\subsection{Possible World Semantics}
 
Kripke semantics~\cite{Kripke63} adds possible worlds to Tarski
models~\cite{Tarski56}. Instead of having a model represent a world, a model now
contains many possible worlds and an {\em accessibility
relationship} $R$ between possible worlds indicates what other
worlds are possible from a given world.  For example, suppose our
model $M$ contains a set of possible worlds $W$ with $w_1$ and
$w_2$ member worlds in $W$. Furthermore, our model will contain an
accessibility relationship $R$ that holds if one world can be
reached from another world.  Let $\Phi$ be a function that assigns
truth $T$ and false $F$ to sentences depending on the conditions
that hold in the model $M$.   Then, the truth condition for the
sentence $\Diamond \alpha$, (read as "It is possible that
$\alpha$" can be written as follows: \begin{eqnarray}
\Phi(\Diamond \alpha, w_1) = T \mbox{ iff }  \exists  w_2
\in W, \mbox{ if } w_1 R w_2 \mbox{ then } \Phi(\alpha, w_2) = T
\end{eqnarray}

This states the sentence ``It is possible that $\alpha$'' is true
in the world $w_1$ if and only if there is some world $w_2$ that is
accessible from $w_1$ and $\alpha$ is true in that world.  In
effect, the truth condition repeats the intuitive meaning of the
sentence but in a formal description that relates the sentence to
the conditions of the model that represents the structure of the
universe (which,  in this case, consists of many possible worlds).

\subsection{Semantics of Temporal Logic}

More informally, A.N. Prior~\cite{Prior67} investigated the
semantics tense operators, like "It will be the case that
$\alpha$", $F\alpha$,  before Kripke. But, it turned out that one
can reinterpret the accessibility relationship in Kripke semantics
in many ways. One way is to interpret the accessibility
relationship as a temporal ordering relationship. In the case of
linear time $R$ is just $<$, the less-than relationship between
time points. This can then be used to give a formal semantics to
the Prior tense operators as well as others. Summing up, in
Kripke's extension to Tarski semantics a model is a tuple $<W, R,
O, P^n, \Phi>$ where $W$ is a set of {\em possible worlds} and $R$
is an accessibility relationship between possible worlds.  The
subportion $<W, R, O, P^n>$ is called a {\em model structure} as
it is somewhat  independent of the particular truth conditions
function $\Phi$ of the complete model.

\subsection{Limitations of Kripke Possible World Semantics}

Note, there are no agents in a Kripke model.  A further subtlety is
that any sophisticated object language contains both tense
operators and modal operators. For example, one can speak of what
was possible or what will be necessary.  Such mixed modal systems
that contain several types of modal operators are becoming
increasingly common and being used in the formal specification and
verification of distributed and multi-agent systems.  They were
first investigated in~\cite{Werner74}.  The main results were
republished and are available in
\cite{WernerMLGames,WernerTensedML}. Both  modal operators with
temporal indices, as well as, logics with both tenses and modal
operators were investigated.

Modal operators with temporal indices allow one to express things
like $\Box_{t_1} \alpha(t_2) \rightarrow \Box_{t_3} \alpha(t_2)$
where $t_1 \leq t_3$.   This states that if it is necessary at time
$t_1$ that sentence $\alpha$ holds at time $t_2$ then it will be
necessary at any later time $t_3$ that $\alpha$ holds at time
$t_2$. Tensed modal systems allow a similar sentence: $P\Box\alpha
\rightarrow \Box P \Box\alpha$ where $P$ is the past operator for
"It was the case that" and $\Box$ is the necessity operator.  The
formula states "If it was necessary that $\alpha$ then it is
necessary that was necessary that $\alpha$." Since the above two
axioms hold in models were the information of an agent has does not
decrease over time, I called these axioms the NDI axioms
(Nondiminishing Information axioms). These were part of a
systematic investigation of logics for agents acting in a changing
open world of choices given only partial information about the
world.
 
\subsection{Problems with Branching Time} 

The semantics for these multi-modal systems become more complex.
Now we need to include not just possible worlds, but also time.  In
a Kripke model the worlds are static.  However, as we noted above,
the accessibility relation $R$ can be interpreted as a temporal
relationship.  Thus, if $R$ is linear we can interpret it as a
linear time ordering.  Particular temporal axioms then hold.   Many
authors attempt to use Prior's original solution to modeling
choices in a  temporal framework by using his notion of {\em
branching time}.  In  branching time the temporal relationship is
not linear but allows  branches into the future (given the
appropriate assumptions about the  temporal relation $R$.)  We then
get a tree-like temporal structure  where each path in the tree
from start to finish (given a finite tree)  represents a possible
history of the world.
 
There are difficulties with branching time, as Prior himself 
recognized.  First, it is very counter-intuitive to the normal view of 
time as linear.  Second, times in distinct possible histories are 
incomparable.   Yet, we regularly reason, both hypothetically and 
counterfactually, about possibilities that involve comparing times in 
distinct possible futures and pasts.  Third, a more subtle, but central 
point is that the semantic representation of uncertainty about time 
becomes difficult.  Fourth, the computation of temporal comparisons 
becomes more complex because one has to check the partial ordering 
relationships versus checking direct linear numerical relationships.  
Fifth, the use of time to represent indeterminism forces an 
identification of possibilities with time points.  This is not only an 
ontological confusion but also a category mistake that jeopardizes the 
semantics of both time and world states.  For these and other reasons, 
I rejected the use of branching time in favor of linear time with 
indeterministic worlds.   

\subsection{Dangers of Whimsical Accessibility Relations}

A scientist who blindly adheres to Kripke semantics may try to
introduce a separate accessibility relation for each new operator
he comes across without trying to understand the function and
structure of those relations We have seen how Kripke introduced an
accessibility relation for the necessity and possibility operators,
and we have noted the use of an accessibility relation for
representing branching time.

One might also be tempted to postulate an accessibility relation
for the 'believes that' operator and the 'intends that' operator.
Unfortunately, in doing this all hope of actually understanding the
relationship between such operators, e.g., 'believes that' and
'intends that', becomes problematic.  For, without giving an
accessibility relation properties, one accessibility relation is no
better than and no different from any other accessibility relation
holding between possible worlds.  Discovering those properties of
accessibility relations is one of the key problems in developing
formal semantics. Furthermore, in postulating that an operator's
semantics is given by an accessibility relation between possible
worlds one may actually no longer be making any sense.  To say that
'$A$ intends that $\alpha$' holds if the 'intends that'
accessibility relation holds for some set of worlds fails to relate
the logic of intends to plans.  It, instead, is based on a false and
blind analogy with the 'necessary that'  and 'possible that'
operators.

For this reason I try to integrate the concepts of time,
information, ability and intention as constitutive parts of the
semantics and pragmatics.  I do so, not by blindly adding
accessibility relations between possible worlds, but, instead, I
construct the specific formal semantic objects that are appropriate
to the meaning of the operator.  And, providing this meaning, is
after all, the purpose of a semantics. Because of this, I will, for
example, define 'can' and the 'intends that' operators in terms of
plans and strategies and structures constructed from or derived from
such plans and strategies. We, thereby, are more likely to get at
the actual meaning of these operators.

The failure to provide structure to the intends and belief
accessibility relations by connecting the semantics with the actual
meaning makes, for example, the relationship of intentions with
abilities very difficult if not impossible to see.  Just as
difficult to formulate is the relationship of information (beliefs)
with intentions, as well as, with abilities. Furthermore, our
semantics allows the investigation of what happens in cases of
partial information about the world, as well as the more difficult
topic of partial information about plans.  The reason we are able
to formalize these relationships is because the semantics  I 
developed, in contradistinction to other semantics, integrates
dynamic plans and dynamic information states into a temporal
possible worlds semantics.

\section{Dynamic Possible World Semantics}

\subsection{Avoiding Branching Time}

How we avoid using branching time? We simply refuse to identify
time points with world states or situations.  Instead, we associate
world states with times, both having equal ontological status but
being of different categories of entities.  Thus, a world history
becomes a function from the set of times to the set of possible
states.  More formally, let $T$ be a set of times linearly ordered
by a relation $<$. Let $\Sigma$ be the set of all possible states
of the system. Then a {\em history} $H$ is a function from $T$ to
$\Sigma$.  Let $H_t$ represent the state of the world $H$ at time
$t$ in $T$.  That is, $H_t = H(t)$.  Let $H^t$ be the {\em partial
history} of the world $H$ up to and including time $t$.
Intuitively, $H^t$ is the history of the world up to the time $t$;
it is not defined after time $t$.  But prior to and including time
$t$ the partial history $H^t$ behaves just like $H$. This is because
they are identical prior to time $t$, i.e.,  $H_{t_0} = H^{t}_{t_0}$
for all times $t_0 \leq t$.   A branch into the future at time $t$
is consists of two histories $H$ and $K$ such that $H$ and $K$ are
identical up to the time $t$ but diverge thereafter.  We then say
the worlds $H$ and $K$ are {\em backwards identical} at time $t$.
Let $\Omega$ be the set of all possible histories $H$ that are
allowed by the system, universe, or model in question. We now are
able to represent indeterministic situations without using
branching time.

\subsection{Dynamic Accessibility Relationships} 

In Kripke semantics the accessibility relationship is static, it
does not change with time.  Yet, in the real world possibilities
change with time and intuitively the worlds that are accessible even
from the same world can change time.  In chess, you have the
classic example that you cannot castle the king if the king has been
moved before, even if the king has been moved back to its original
position.  Thus, what is possible after the king has moved is no
longer possible in the future even if the state of the chess game is
identical to one where one could otherwise castle.    More
generally, it is helpful to consider relationships that change with
time when investigating the dynamic situations that agents find
themselves in, especially when the agents no longer have perfect
information about the state of the world.
 
A world history, much like the world line  in physics, can be
viewed as a world that changes in time.  So, analogous to world
histories, a changing relationship can be defined as a function
from times to possible static relationships.   The result is a
dynamic relation that changes with time.  In particular, we can
generalize the Kripke accessibility relationship to one that
changes with time. More formally, a dynamic accessibility relation
${\cal R}$ is a function from times $T$ to relations on $\Omega$.
Let ${\cal R}^t$ be the value of the dynamic accessibility relation
${\cal R}$ at time $t$.  Two histories $H$ and $K$ are related by
${\cal R}$ at time $t$ if $H {\cal R}^t K$. Recall a normal static
relation is just a set of ordered pairs.  On this view, two objects
$a$, $b$ stand in a static relationship $R$ if the ordered pair
$(a, b)$ is a member of $R$.  A {\em dynamic relation} ${\cal R}$
is a changing set of ordered pairs. So the value of a dynamic
relationship at a time $t$, in symbols ${\cal R}^t$, is a static
set of ordered pairs. Hence, $H$ is related to $K$ by ${\cal R}^t$
if and only if the ordered pair $(H, K)$ is a member of ${\cal
R}^t$ which is what $H {\cal R}^t K$ says.
 
\subsection{Models with Dynamic Worlds in Changing Relationships}

So far so good.  We now have changing indeterministic worlds that
can be in changing relationships to one another.  Time is still
linear. We have extended the Kripke model significantly.  Given a
temporal modal language $L$ (still unspecified) a  {\em semantics}
for temporal modal logic consists of  a dynamic open model
structure 
\begin{eqnarray}  \mbox{{\cal MS}} = ((T, <), \Sigma,
\Omega, {\cal R}, O, P^n)   \end{eqnarray}   together with an
evaluation function $\Phi$ where:
 
\begin{enumerate} 
 
\item $T$ is a set of times linearly ordered by a temporal ordering 
relation $<$. 
  
\item $\Sigma$ is a set of possible states. 
   
\item $\Omega$ is a set of dynamic world histories.  
 
\item ${\cal R}$ is a dynamic accessibility relation between the
elements of $\Omega$.

\item  $O$ is a set of objects. 

\item $P^n$ is a set of properties and relations of order $n = 0$
to some finite $N$.

\item $\Phi$ is an evaluation function giving the truth (and
satisfiability)conditions for the given temporal modal language
$L$.

\end{enumerate}

\subsection{Dynamic Semantics for Temporal Modal Logics}

We already have a quite powerful semantics.  We can investigate
the dynamic modal logics of agents in changing circumstances.  The
logics can be temporally indexed modal operators, like $\Box_t$.
Or, we have tenses combined with modal operators, as in $\neg F \neg
\Box \alpha$ where $\neg$ is the negation operator "It is not the
case that", $F$ is the future operator "It will be the case that"
and $\Box$ is the necessity operator "It is necessary that".  Thus,
literally this formula states "It is not the case that it will be
the case that it is not the case that it is necessary that
$\alpha$".  Equivalently and more succinctly this says "It will
always be necessary that $\alpha$."

The truth condition for
$\Box_t$ is as follows: 
\begin{eqnarray} 
\Phi( \Box_t \alpha(t'), H) = T \mbox{ iff for all } K \in \Omega,
H {\cal R}^t K, \Phi( \alpha(t'), K ) = T \end{eqnarray} 

Translating into English, this says that "It is necessary at time
$t$ that $\alpha$ is true ($T$) at time $t'$" is true in the world
history $H$ if and only if for all world histories $K$ such that $K$
is possible in the history $H$ at time $t$ relative to the dynamic
conditions ${\cal R}^t$. The time-dependent relationship ${\cal
R}^t$ expresses conditions such as information conditions in the
world $H$ that determine what is possible at that point in time.

The analogous truth condition for the tense modal language is:
\begin{eqnarray} \Phi( \Box \alpha, H, t) = T \mbox{ iff for all } 
K \in \Omega, H {\cal R}^t K, \Phi( \alpha, K, t ) = T
\end{eqnarray}   

Intuitively, this is like the Kripke interpretation except now
what is possible and necessary depends on the time $t$ and
changing accessibility relation ${\cal R}$ holding over changing
worlds $H$ and $K$.  The properties of the accessibility
relationship determine the kind of modal axioms that hold
\cite{HughesCresswell}, except now those properties include
temporal properties of changing relationships. We now look at what
is possible in a changing world at a given point in time. Temporal
relationships are evaluated distinctly from modal relationships in
the case of tense operators. If, for example, $\alpha$ is a tense
formula of the form $F\beta$ then $\Phi( \Box F\beta ), H, t)$
holds if and only if for all $K$, $H {\cal R}^t K$, $\Phi( F\beta,
K, t )$ holds. And this holds if and only if for each $K$, $H
{\cal R}^t K$, there is a time $t'$ where $t < t'$ and $\Phi(
\beta, K, t' )$ holds. Hence,  $\Box F\beta$ holds in world $H$ at
time  $t$ if $\beta$ holds in every possible future $K$ at some
future time $t'$.  In other words, $\beta$ is inevitable in the
world $H$ at time $t$, but the exact time at which $\beta$ will
hold is not specified.

\subsection{Properties of the Dynamic Accessibility Relation}

Interestingly, some general axioms hold depending on the dynamic
properties of the dynamic accessibility relation.  Two of the most
important properties are backwards identity and backwards
consistency.  Later we will also see that the information
conditions of an agent generate a dynamic, information relation
that is also a dynamic, accessibility relation.  There is an
interesting relation between the information conditions of an agent
and the properties of the dynamic, accessibility  relation.

\subsubsection{Backwards Identical Dynamic Relations}

A dynamic relation ${\cal R}$ is {\em backwards identical}  if  for
all worlds $H$ and $K$ in $\Omega$ and for all times $t$  in $T$,
$H {\cal R}^t K$ if and only if  for all times $t_0 \leq t, H_t =
K_t$. This just says the two worlds are identical prior to and
including time $t$.  We will see below that when an agent in a
system has perfect information, in the sense of knowing his complete
past and present states, then the accessibility relation generated
by the information conditions is backwards identical.
 
\subsubsection{Backwards Consistent Dynamic Relations}

A dynamic relation ${\cal R}$ has {\em non-diminishing information}
or{\em satisfies the NDI-condition} if  for all times $t$ in $T$,
for all times $t'$ where $t \leq t'$, if $H {\cal R}^{t'} K$ then
$H {\cal R}^{t} K$. This says the dynamic relation ${\cal R}$
cannot grow larger in the future. We will see later why this
implies information is not lost.   Note, this does not say that more
options cannot be available in the future. Indeed, the opposite can
be true. Rather it only says the world {\em histories} that are
related cannot increase under the backwards consistency condition.
Note, also, that the backwards consistency condition is implied by
the backwards identity condition. Later we will see the
informational analog of this, namely, perfect information implies
non-diminishing information.

\section{Situation Semantics and Pragmatics}

\subsection{From Montague to Situation Semantics}

In the seventies Montague semantics was king of the hill.  It was
a systematic extension and application of the Tarski, Kripke
semantics to give a semantic analysis of language (see
Montague~\cite{Montague76}.)  Because of its extensive use of
possible possible worlds and accusations of computational
intractability, it fell out of favor in the eighties.  Taking its
place was the challenger, situation semantics, developed by
Barwise and Perry~\cite{BarwisePerry83}. It is this challenger
that held the world of computer science captive for almost a
decade.  It was the semantics embraced by the "Fifth Generation
Project" in Japan, the project that was to produce a thinking
expert machine by the end of the 80's.

\subsection{Motivations for Situation Semantics}

Besides many counter-intuitive complex structures in Montague
semantics, one motivation Barwise and Perry may have had was the
following idea:  Why use possible worlds if we can build up
semantic structures directly in terms of world objects, properties
and relations.  Meaning then becomes a direct relation of language
with the world and is not mediated by way of truth conditions and
possible worlds.  More deeply, one may view situation semantics as
an attempt to represent information positively, and not
extensionally, about the world. Instead of looking at the set of
possible worlds where Bill sits on a horse and concluding the
sentence "Bill sits on a horse", is known with certainty if Bill
sits on a horse in all accessible possible worlds, one just
represents it as a direct situation, e.g., {\bf sits-on(Bill,
Horse)}. For simplicity, I have left out time and other details.
The basic idea is there though,  represent meaning as a set of
situations that a sentence refers to.  This avoids a nondenumerable
set of possible worlds and makes the representation of semantics in
a computer at least feasible.  The goal is then to map a situation
with each possible sentence of the language, instead of the earlier
truth conditions.

\subsection{Problems with Situation Semantics}

Unfortunately, there are difficulties with the approach.  They
involve exactly what situation semantics tried to avoid, namely,
possibilities. Consider, for example, negation. What does "Bill is
not sitting on a horse" mean?  It is not a unique situation where
Bill is not sitting on a horse, for there are infinitely many such
possible situations.  Also disjunctions were a problem.  For, there
again we have possibilities, e.g., "Bill is sitting on a horse or
Bill is in the kitchen".  Quantifiers are also difficult.  The
result was that Barwise and Perry were forced to include sets of
possible situations into the semantics.  While situation semantics
is still interesting, it began to look suspiciously like a more
encompassing possible world semantics (possible worlds were just
complete situations).  A further, more subtle, but very important
problem is that the attempt to represent all information positively
without the use of possibilities is not as powerful as the possible
states approach because there will always be information states
that are not representable as positive situations.  Yet, these
possible information states are, of course, representable by sets
of possibilities in the information as possibilities approach.

The above problems are faced by any theory meaning that tries to
represent information positively instead of extensionally in terms
of possibilities.  Yet another problem with situation semantics is
that it is just about situations, it fails to include the subject or
agent and his mental life into the theory of meaning.  However, to
understand social cooperation by use of linguistic interactions, we
need a theory of representations of intentions.  And, we need  a
theory of how language transforms such representations.  For this
we need a theory of pragmatics.

Still the idea that referential meaning and simple propositional
content can be positively or directly represented is I believe a
good one and has a place as a component in a general theory of
meaning.

\subsection{Situations by a Different Route}

Around the late seventies at about the same time Barwise was
developing situation semantics, I was intrigued with speech acts,
like commands and requests. Since such speech acts are not true or
false, giving commands and requests a classical truth-conditional
semantics appeared questionable and inappropriate. I realized that
the theory of meaning for commands, requests, and questions was
going to be fundamentally different from the semantics of true and
false sentences that researchers had concentrated on previously.
While assertions focus on the world, commands and requests focus on
controlling or influencing the actions of another agent.  My
hypothesis was that they control or influence another agent by
changing the intentions of the agent, since the intentions were plan
like states that guided the agent. I developed the concept of
intentional state.  These states were to be like information states
(the formal theory of which I had already developed) in that they
must be dynamically changeable by communication with other agents.
Taking my theory of information states and my theory of
communication for information states as a basis, the next key step
for me was to develop the formal theory of intentional states.

The development of such a theory involved keeping in mind the
requirements of a theory of communication and cooperation. The
theory of intentional states was thus  developed simultaneously
with the pragmatic theory of meaning and communication for
such intentional states (outlined above). I decided to test
the developing theory on larger and larger fragments of English
including directives, requests, and informatives as well as
prepositions, and tensed verbs.  In the attempt to provide a
semantics for such English sentences, it became evident that it was
useful to give verbs and basic sentences a referential meaning that
directly refers to objects having properties and standing in various
relationships with one another. My idea was to give a basic
referential semantics to atomic sentences in terms of what I called
{\em relational structures} (including relational space time
structures or events).

\subsection{Deep Referential Situation Semantics and Pragmatics}

Later, while I was investigating a large but relatively flat
fragment of English (English for foreigners), I saw quickly that
negation, disjunctions, quantifiers, etc. were going to cause
difficulties if one tried to give a direct relational structure
semantics to all sentences of English.  Since, however, language
does refer to and link up with the world, I needed to formalize the
propositional or referential content of sentences.  To solve the
problem I divided the theory of meaning into two layers: The
first layer, which I called the {\em deep semantics}, was a
semantics referring to situation-like, relational space-time
structures.  A second layer, the {\em pragmatics}, allowed the
interpretation of more complex sentences by describing their effect
on the information and intentional states of the agent.  The deep
semantics was actually developed out of the need to give a full
compositional pragmatics for English like fragments.  The
compositional pragmatics arose out of the attempt to understand
communication.  And the theory of communication arose out of the
attempt to understand cooperation between groups of agents.  This
methodology of having one discipline constrain another is quite
useful for the development of ideas.

\section{Information and Uncertainty}

If we consider planning and action from the perspective of an agent
$A$ who is acting in not just a physical environment $\Omega$, but
also, in a social environment of other groups $G$ of agents, then
the state of uncertainty, or what the agent knows and does not
know, can be very complex. Fundamental to a clearer understanding
the complex relationships of uncertainty to action and intention,
is the rigorous description of these kinds of uncertainty. 

But, before we do this, it is important to note that information
and uncertainty are two sides of the same coin. Each type of
uncertainty (e.g., state, temporal, intention, plan uncertainty),
has a corresponding type of information state (e.g., state
information, temporal information, intentional state, plan state)
that positively describes the uncertainty in terms of information
the agent possesses.  

\subsection{Imperfect and Perfect Information}

We will say an agent is in a {\em state of imperfect information}
or, simply, the agent has {\em imperfect information} about the
world, the consequences of an action, his own plans or the plans of
others, or the consequences of a plan or set of plans, if that agent
has state uncertainty, action uncertainty, plan uncertainty, or plan
consequence uncertainty, respectively.  Otherwise, the agent has
{\em perfect information} about state, action consequences, or
plans.

An agent in a state of imperfect information may either have some
information or no information.  To say an agent has {\em total
uncertainty} is equivalent to saying the agent has no information.
If the agent has perfect information of some kind, it is equivalent
to saying the agent has no uncertainty of the kind in question.

\subsection{Relating Information States and States of Uncertainty}

Thus, we get the following equivalencies:  {\em No Information
$\Longleftrightarrow$ Total Uncertainty, Partial Information
$\Longleftrightarrow$ Partial Uncertainty, Perfect Information
$\Longleftrightarrow$ No Uncertainty}. A minimal requirement of any
theory of information and uncertainty is that it be able to define
and distinguish perfect information (no uncertainty), from
imperfect information (partial uncertainty), and no information
(total uncertainty). In what follows we will talk either of
uncertainty or information depending on which side of the coin we
want to emphasize.

In the following sections, we will formally define these different
types of information and uncertainty by defining the different
categories of states of information and uncertainty of agents. This
will make possible the investigation of the precise
interrelationships between these types of uncertainty and
information.  We start with describing state information and
uncertainty about state and, later, we will give a formal
characterization of strategic uncertainty (information about
intentions, strategies and plans). This will then give formal
substance to the more abstract summary of the {\sf ICE} agent
architecture described above.

\section{State Uncertainty}

Agents acting in a world will generally not have full information
about the state of the world.  The semantics of Tarski and, also,
Kripke semantics make no provisions for partial information about
the world. This is because those semantics were never intended to
deal with problems of agent cooperation and communication.  The
dynamic semantics provides a framework for describing changing
accessibility relationships in a dynamic changing world of
possibilities. However, the exact interpretation of the properties
of a dynamic accessibility relation is left open.  We now extend
the semantic framework to include partial states of information
about the world.  

\subsection{Possible Situations Over Time}

First, we need some preliminary concepts. Let $\Psi$ be the set of
{\em macro time instants} ordered by a relation $<$. Let $TP$ be
the set of time periods over $\Psi$. We will use $t$ to represent
instants and $\tau$  to represent time periods. Let  $s$  be a
situation at a given instant. A situation is a partial
representation of the state of the world~\cite{BarwisePerry83}. Let
$Sit$ be the set of all possible situations. An {\em event} $e$ is a
partial function from the set of times into the set of possible
situations, $e : \Psi \rightarrow  Sit$.  Let EVENTS be the set of
all possible events.

Let a {\em world state} $\sigma$ be a total description of the
state of the world at a given instant. Hence a world state will be
a totally defined situation. Let $\Sigma$ be a set of states. A
{\em (total) history} $H$ is a function from $\Psi$ into $\Sigma$.
A history is thus a series of complete situations over the time
period $\Psi$.  We will also refer to histories as possible worlds,
possible past and futures, or as world lines.  Let $\Omega$ be the
set of all total histories each indexed by $\Psi$. If $H  \in
\Omega$ , then let $H_{t}$ be the value of the function $H$ at $t
\in  \Psi$. $H_{t}$ represents the state of the world $H$ at time
$t$. $\Omega$ then represents the set of all possible changes the
system can undergo given the constraint that these changes are
allowed by the rules or laws governing the system.  A given history
$H$ {\em realizes} an event $e$ over period $\tau \in TP$ iff
$Domain(e) = \tau$  and $\forall t  \in Domain (e), e_{t} \subseteq
H_{t}$.

A  {\em partial history} $H^{t}$ of $H$ is a partial function from
$\Psi$ into $\Sigma$ such that  $H^{t}_{t_0} = H_{t_0}$ if $t_{0}
\leq t$ and undefined if $t < t_{0}$ .  $V(\Omega)$ represents the
class of all partial histories in $\Omega$.	These $H^{t} \in
V(\Omega)$ are also called the {\em vertices} of $\Omega$.
 
\subsection{Information States}

State information is that which reduces state uncertainty. By state
information we mean information about the state of the world.  For
example, information about the state of the cards in a game of
cards or the location of a robot in a room at a particular point in
time is state information. We distinguish it from strategic
information which is information about agent plans and intentions.

An {\em information set} or {\em I-set} on $\Omega$ is a nonempty
class $I$ of partial histories in  $\Omega$ such that for any $t$,
$t' \in \Psi$ and any $H \in \Omega$, if $H^{t} \in  I$ and $H^{t'}
\in I$ then $t = t'$.  We will use the letter $I$, $J$ with and
without superscripts to denote information sets.  For a given agent
$A$ the information set  $I_{A}$  represents the information state
of the agent at some time in a given world. If $I$ is any
information set on $\Omega$, let  $I^{*}$  be the class of worlds
intersected by $I$.  $I^{*}$ is the set of histories allowed by the
information $I$.  
\begin{eqnarray}
I^{*} =_{df}  \{H: H \in \Omega \mbox{ and there is a }
t  \in \Psi \mbox{ such that } H^{t}  \in  I\}
\end{eqnarray}  
With each information
set $I$ we associate a set of {\em alternatives} $Alt (I)$.
Alternatives are the choices available to the agent given the
information $I$. These alternatives are generated from the effect
of microactions on the present state (see the section on
simultaneous choices).  Note, the alternatives leaving an
information set are different from the possibilities $H^t$ within
the information set.

\subsection{Information and Possibility} 

Information sets have an interesting property, which we call the
{\em Entropy Principle}:  The information available to an agent is
inversely related to the number of possibilities in the
information set.  A corollary principle, which we call the {\em
Principle of Possibility Reduction PPR}, is the following: The
more information becomes available the fewer the possibilities.
This principle also holds for information sets. To illustrate this
principle consider an example: Mary, who is playing cards with Joe,
knows Joe has the Queen of Hearts $(Q\heartsuit)$, but she does not
know if Joe has the King of Hearts $(K\heartsuit)$ or the Jack of
Spades $(J\spadesuit)$. Mary's information state $I_{1}  =
\{\sigma_{1} , \sigma_2  \}$  is one where she does not know if
Joe's hand is in state $\sigma_{1} = ~ <Q\heartsuit, J\spadesuit>$
or in state $\sigma_2   = ~ <Q\heartsuit, K\heartsuit>$.  The
information that Joe has the King of Hearts reduces the information
set $I_{1}$ (by the  possibility reduction principle) to the
information state $I_2 = \{\sigma_2\}$ where Mary knows with
certainty (has perfect information) that Joe has the hand $\sigma_2
= ~  <Q\heartsuit, K\heartsuit>$.   Note, here information sets
consist of states $\sigma$ which are partial histories that are one
instant of time in length (see~\cite{WernerUniView} for further
discussion about these and other principles).  Starting from very
similar intuitions, a complementary approach to state information in
planning  is taken by Steel~\cite{Steel91}.

\subsection{Temporal Uncertainty}

\subsubsection{Imperfect About Time} 

Let   $Time(I_{A})  =  \tau$ be the time period specified by the
information state  $I$.  It is the temporal information given by
$I$.  An information set $I$ is {\em straight} if for any $H$, $K
\in  \Omega$, if $H^{t}  \in  I $ and $K^{t'} \in I$ then $t = t'$.
An information set that is not straight will be said to be {\em
slanted}.    Straight information sets give perfect temporal
information about the $Time(I)$ because any two partial histories in
$I$ pick out the same unique present time $t$. Slanted information
sets give imperfect information about time. Formally, this means
that for two partial histories $H^t$ and $K^{t'}$ that are
possibilities in $I$, their present times $t$ and $t'$ may differ.
This means that given this information $I$, we do not know if the
present time is $t$ or $t'$.

\subsubsection{Perfect Temporal Information}

An information set is {\em thin} if $H^{t} \in  I$ and $H^{t'}  \in
I$ then $t = t'$. Intuitively, this means that the for a given
history $H$ the agent has no uncertainty about time.  In effect,
perfect information about the history and state of the world gives
the agent perfect information about time. The general restriction on
information sets is that they are thin.  But, one could imagine
situations where the agent is uncertain about the time of the world
even though he knows exactly what has occurred up to his range of
uncertainty.  He just does not know if he is in the future or in the
past relative to his temporal uncertainty. This situation may hold
if the world can have two successive states that are absolutely
identical. The only difference is the time.  An example might be an
agent who sits passively in a room without a clock and who
loses track of the time.  Note, the properties thin and slanted are
independent.  This means that an information set can be thin and
slanted, not thin and slanted, thin and not slanted (straight), or
not thin and not slanted (straight).   

\subsection{Information	Conditions}

The following is used to represent all possible information
conditions of a given agent:

An {\em information ensemble} for $\Omega$ is a class  $\Xi$  of
information sets  on $\Omega$  such that the following conditions
hold:

\begin{enumerate}

\item For any $H^{t}  \in  V(\Omega)$, there is an information set
$I \in \Xi$ such that $H^{t}  \in  I$.

\item For any $I$, $J \in \Xi$, if $I \neq J$ then $I \cap J =
\Lambda$, the null set.

\end{enumerate}

We will use the symbols $\Xi$, $\Xi'$ to denote information
ensembles. 

An information ensemble is a complete partition of the vertices in
$\Omega$. It follows immediately from the definition, that if  $\Xi$
is any ensemble for  $\Omega$, then for any partial history $H^{t}
\in V(\Omega)$, there is a unique information set $I$ in $\Xi$ such
that $H^{t}  \in  I$. Let this unique information set be denoted by
$I(H^{t})$. Let the {\em information history} $I(H)$ be a function
from $\Psi$ to $\Xi$ where $I(H)(t) =  I(H)_{t} =_{df} I(H^{t})$ for
each $t  \in  \Psi$.

An information set $I(H^{t})$ represents the information available
to an agent in the world $H$ at time $t$. $I(H)$ represents the
changing information conditions throughout the history $H$.  An
information ensemble $\Xi$ then gives the information conditions for
an observer for all possible developments of the given system. We
interpret an ensemble as relativized to an observer-agent. The
information conditions for $n$ agents are then given by $n$
distinct information ensembles $\Xi_{1}, \ldots , \Xi_{n}$.

\subsection{Nondiminishing Information}

Let $\Xi$ be any ensemble for $\Omega$ , then there is {\em
nondiminishing information} in $\Xi$ if for all $H \in \Omega$ and
all $t, t' \in \Psi$, if $t \leq t'$ then $I(H^{t'})^{*} \subseteq
I(H^{t})^{*}$.  If there is nondiminishing information in $\Xi$,
we call $\Xi$ an {\em NDI-ensemble}.  An increase in size of an
information set $I(H)$ represents a decrease in available
information. For, an increase in the size of $I(H)$ is an increase
in the possibilities.  By insuring that the information set $I(H)$
does not expand in the future, the condition of nondiminishing
information guarantees that no information is lost once it is
stored.  A different way of putting this is that no new, unforseen 
possibilities may arise.

\subsection{Perfect State Information}

The condition of nondiminishing information is the weakest
information condition one can place on an ensemble and still have
no loss of stored information. We now define the strongest
condition one can place on the information.

Let $\Xi$ be any ensemble for $\Omega$, then there is {\em perfect
state information} in $\Xi$ if for each $I \in  \Xi$, $I$ is a
unit set, i.e., if $\forall v_{1}, v_{2} \in V(\Omega)$, if $v_{1}
\in I$ and $v_{2}  \in  I$ then $v_{1} = v_{2}$.

A decrease in the size of the information set represents an
increase of information. When $I(H^{t})$ is a unit set for some $H
\in \Omega$, the observer has total knowledge of the history of
the system up to and including the time $t$.  Chess and checkers
are examples of games with perfect information; most card games
are games of imperfect information.

\subsection{Definition of an Information Relation}

The information conditions of an agent generate a Kripke like
accessibility relation with the difference that it is a dynamic
relation that varies with time and information.  Given an ensemble
$\Xi$ for $\Omega$, the variable {\em information relation}
$\Im^{\Xi}$ generated by $\Xi$ is defined as follows: For any $H, K
\in \Omega$ and any $t \in \Psi, H \Im^{\Xi}_{t}K$ if and only if
$K \in I(H^{t})^{*}$. 

Intuitively,  $\Im^{\Xi}$  is a relation that varies with
time. $H \Im^{\Xi}_{t}K$  says that $K$ is a possible outcome of
$H$ given the information available in $H$ at time $t$. $K$ is an
associated world of $H$, but need not be accessible  to $H$ in the
sense of being an {\em actual possibility}. $K$ may only appear to
be possible because of the limited information available in $H$.
We can abstract away from the reference to the information ensemble
$\Xi$ and state information conditions as properties of a temporal
accessibility relation~\cite{WernerMLGames}.

\section{Information, Semantics, and Temporal Logic}

\subsection{Information Based Semantics}

The dynamic information relation $\Im^{\Xi}$ can be used to define
{\em information based semantics} for both tensed and temporally
indexed modal logics.  To do this we simply replace, or better,
instantiate the more abstract dynamic accessibility relation ${\cal
R}^t$ with the information relation  $\Im^{\Xi}_t$.	 For
the temporally indexed necessity-information operator we have:

\begin{eqnarray} 
\Phi( \Box_t \alpha(t'), H) = T \mbox{ iff for all } K \in \Omega,
H \Im^{\Xi}_t K, \Phi( \alpha(t'), K ) = T \end{eqnarray} 

For the tensed modal version of the necessity-information operator
we have: 

\begin{eqnarray} 
\Phi( \Box \alpha, t, H) = T \mbox{ iff for all } K \in \Omega, H
\Im^{\Xi}_t K, \Phi( \alpha, t, K ) = T 
\end{eqnarray}
 
Note, how the time index has been moved from the semantic truth
conditions to the formulas of the object language.

\subsection{Relating Logic, Information and Time}

The surprising discovery made
in~\cite{Werner74}(see~\cite{WernerMLGames,WernerTensedML} was that
the properties of the dynamic, accessibility relation have direct
correlates in properties of the information conditions holding in
open multi-agent systems. Thus, the axioms that hold can be
interpreted as corresponding to abstract properties of
time-dependent accessibility relations, or as reflecting types of
information conditions of agents.

Let me put this last, important point another way. We have seen
that the information conditions of an agent generate an information
relative, dynamic, accessibility relations between possible
histories. And, this information relation was used to construct a
semantics. Earlier we saw that dynamic accessibility relations can
have various properties. These properties determine what temporal
modal axioms hold of a system. But, since these properties
correspond to information conditions, the axioms are also
informational principles. 

\subsection{Temporal, Information Based, Modal Logics}

We now look at some important axioms that hold for
various information conditions. 
We will look at both temporally indexed versions and tensed
versions of the axioms.  The mixed tensed modal systems resulting
from axioms that correlate both tenses and modal operators, are
of special interest because they cannot be generated by simply
combining standard modal and tense logics.

\subsubsection{Axioms for Nondiminishing Information}

The following schemata of a temporally indexed modal language is
valid in every model for the language when the dynamic,
accessibility relation is backwards consistent.  But, since we
have the theorem that any system with nondiminishing information
has a dynamic accessibility relation that is backwards consistent,
it follows that these axioms also hold for any system with the
nondiminishing information condition holding.

For {\em NDI axiom for temporally indexed formulas}:   For all $t$,
$t'$, $t''$where $t \leq t'$, 
\begin{eqnarray} \Box_t \alpha(t'') \rightarrow
\Box_{t'}\alpha(t'')
\end{eqnarray} 

The axioms says if it is necessary at time $t'$ that $\alpha$
holds at time $t''$ then it is necessary at time $t$ where $t \leq
t'$ that $\alpha$ holds at time $t''$.  In other words, once an
assertion is known then it continues to hold.  This is just the
condition that information is not lost or forgotten; it does not
diminish.  Note, though, that $\alpha$ refers to a specific time
$t''$.

{\em Nondiminishing Information Axiom (Tensed Version)} for $P$ and
$\Box$: 
\begin{eqnarray} 
P\Box_{A}\alpha  \rightarrow \Box_{A}P\alpha
\end{eqnarray} 

The axiom reads "If it was necessary that $\alpha$ then it is
necessary  that $\alpha$ was the case."  Or "If agent $A$ had the
information that  $\alpha$  then $A$ has the information that
$\alpha$ was the case".  Again, it holds only for the case where
the information is nondiminishing~\cite{WernerMLGames,Werner74}.  Also there are
special conditions on $\alpha$.

Analogous to temporally indexed modal axioms
\cite{WernerMLGames,Werner74}, the axioms of tensed modal logics
can be interpreted as informational principles that give
constraints on the information states of an agent.  For example,
the above tensed modal axiom for $P$ and $\Box$ is a partial
monotonicity constraint. Informationally, it asserts that if an
agent has had the information that $\alpha$ in the past, then the
agent will continue to have the information that he had the
information that $\alpha$.  Note, this is a weaker information
condition than complete monotonicity, i.e., no information loss.
For the agent may no longer have the information that $\alpha$, if
$\neg \Box\alpha$ holds.  Yet, the agent will still have the
information that it was the case that $\alpha$, $P\alpha$,  since
the axiom implies $\Box P\alpha$ still holds.

\subsubsection{Axioms for Perfect Information}

There is also a theorem that states that the following axioms
hold if the accessibility relation is backwards identical.
Furthermore, there is a theorem that the information relative
accessibility relation generated by a system with perfect
information conditions is backwards identical.  Therefore, these
axioms also hold for systems with perfect information.  The
temporally indexed axiom that holds under condition of perfect
information is:
 
For {\em PI-axiom for temporally indexed formulas}:  For all $t$,
$t'$ where $t \leq t'$, 
\begin{eqnarray} \alpha(t) \rightarrow \Box_{t'}\alpha(t)
\end{eqnarray} 

As an informational principle it says that if $\alpha(t)$ is true
at time $t$, then it is and will continue to be known.  Note, this
does {\em not} imply that $\alpha(t') \rightarrow
\Box_{t}\alpha(t') $ when  $t \leq t'$.  This says that if
$\alpha$ holds at time $t'$ then it was the case that it was known
at time $t$ that $\alpha$ will hold at $t'$.  This is not
guaranteed by the condition of perfect information, since it
allows uncertainty about the future.

{\em Perfect Information Axiom (Tensed Version)} for P and $\Box$:
\begin{eqnarray} \alpha \rightarrow \Box_{A}\alpha
\end{eqnarray} 

Note, there are restrictions on $\alpha$, in particular, that it
contains no future tenses. This axiom only holds if there is
perfect information about the state of the world.  As an
informational principle it expresses exactly that:  If some fact
$\alpha$ is true of the present or a past state of the world, then
the agent has the information that $\alpha$ holds.  It puts no
constraints on the agent's uncertainty about the future.

Note, too, since we are in a time dependent system, these axioms do
not result in the collapse of the modal operator as it would in a
nontensed system. 

These axioms are important because their conditions of validity
are based on fundamental dynamic, information relationships. These
relationships are universal holding for any multi-agent system with
the requisite properties of their dynamic, accessibility
relationships. For proofs of consistency and completeness theorems
for dynamic modal logics as well as more details on other systems,
their semantics and the relationship to more classical non-dynamic
modal systems see~\cite{Werner74,WernerMLGames,WernerTensedML}.
There it is also shown that von Neumann games satisfy particular
temporal modal logics.

\section{Actions}

\subsection{Action Types}

An action is a type of event.  At the lowest level (in macro-time)
we have {\em token actions} consisting of objects with properties
and standing in various relations that vary in time.  There is a
distinguished object called the {\em agent} who is the initiator of
the action. They are partial situations over time called {\em
events}.  An {\em action type} {\bf a} is then a class of such
token actions.  A token action $a$ is {\em realized} in a world $H$
if and only if $a \subseteq H$. An action type {\bf a} is {\em
realized} in a world $H$ if and only if for some token $a \in {\bf
a}$, $a$ is realized in $H$. With an action and an agent and a
given situation we associate a set of possible {\em consequence
events}. An action type's uncertainty will be the union of all the
possible consequences of each of its possible token actions.

\subsection{Action Uncertainty}

The agent $A$ may be uncertain about the consequences of an action.
This uncertainty may have different causes:  

\subsubsection{Action Uncertainty from Indeterminism}

There may be {\em inherent action uncertainty} where the agent is
uncertain about the consequences of an action because the world
itself is indeterministic. In an indeterministic universe, multiple
possible consequences may be an objective feature of an action.

\subsubsection{Action Uncertainty from Computational Limits}

There is also a {\em subjective action uncertainty} that results
from the lack of information the agent has about the effects of the
action given his computational limitations.  

\subsubsection{Action Uncertainty from State Uncertainty}

However, even if we assume a deterministic universe, and no
subjective uncertainty about an action's consequences for a given
state, the agent may still be uncertain about the action's
consequences since his information state $I$ may not tell the agent
his exact state in the world.  And, since the action's effects
depend on the given state, the agent will not know all the
consequences of his action. This is represented formally by the
fact that an alternative leaving an information set $I$ is a whole
set of individual choices or alternatives, one for each possible
state of the world given the information $I$.

\subsubsection{Multi-Agent Action Uncertainty}

Action uncertainty may also be due to a {\em multi-agent setting}.
In a multi-agent social world a given agent's actions will never
fully determine the next state, because this state is dependent on
what the other agents do as well. Only strategic knowledge about
the other agent's intentions or plan states can reduce this kind of
{\em socially induced action uncertainty}.

Socially induced action uncertainty is a natural part of our
theory. For, in a multi-agent world the state is determined only
after all agents have made some choice.  The {\em do-nothing} or
{\em null choice} being, of course, always one of the options
available to an agent. Formally, any given pure agent strategy
$\pi_A$ will not have a unique outcome in a multi-agent social
world.  Indeed, the more agents there are the less determined is
the result of the strategy $\pi_A$.	Let us look at this phenomenon
in more detail for the case of simultaneous actions by one or more
agents. 

\subsection{A Theory of Simultaneous Choices} 

When agents act they act simultaneously with all the other agents.  
Since the null action is an action, we can assume that each agent 
makes a choice at each moment of time.  

\subsubsection{Why actions cannot be functions from states to 
states}

Traditionally, an action by an agent was viewed as a function from
states to states.  It is viewed as transforming a given state
$\sigma_1$ to a new state $\sigma_2$ that results from applying the
action $a$ to $\sigma_1$.  While this appears to work in the case of
a single agent making a single choice at any given time, this
approach has problems when one considers multi-agents acting
simultaneously or even a single agent performing two actions at the
same time.

More formally, it is common to view an action $a$ as an operator
that takes a state $\sigma$ and transforms the state into a new
state $\sigma' = a(\sigma)$.  As soon as we allow the possibility
of simultaneous actions, an agent's action can no longer be a
function or operator taking one state to a new state.  Why?
Consider two agents $A$ and $B$ where $A$ does action $a$ and $B$
does action $b$. Then given the agents are in state $\sigma$,
$a(\sigma)$ need not equal $b(\sigma)$.  If both operators are
applied to $\sigma$ the result is not a unique state.  Yet,
intuitively, we want to allow two operators to act on a state
simultaneously, but if they do so, the operators are no longer
functions from states to states.  We need a new framework.

\subsubsection{ The Problem of Interference in Simultaneous 
Actions}

The above problem with the traditional theory of action is rooted
in the deeper problem of how to represent action {\em interactions}
(interference or emergence).  Simultaneous actions may influence
and interfere with one another.  An action taken by an agent alone
may have a different effect than if another agent interferes or
helps. For example, in basketball Joe may be perfect at making a
basket if he is playing alone. When, however, Ray is defending the
basket, Joe's performance may suffer.  Similarly, Ray may not be
able to lift the couch himself, but with Joe's help they can lift
it.  In a larger context of many agents, an agent's action,
including communication, may set off a diverse pattern of actions
resulting in a large scale multi-agent event. For example, a
single command such as 'Attack!', when issued by the appropriate
agent may start a war.  We need a theory of simultaneous
actions that allows the description of such interferences and
emergent effects.

\subsubsection{Turn Taking to Solve Agent Action Interference}
 
One way of solving the problems is to have the agents take turns.
First, the action of agent $A$ is applied and then the action of
agent $B$ is applied.    The idea is that agents never really act
at the same time and so cannot interfere with each other.  A good
example of this is the function of traffic lights  and stop signs
to make drivers take turns at intersections.  Unfortunately, and
fortunately, the real world has simultaneous events and activities.
In fact, many actions only have their effect if the actions are
applied simultaneously.  For example, when Dany lifts the heavy
vase she needs to act with both hands at the same time.  She cannot
lift with one hand and then lift with the other hand.  The vase
would never be lifted.  In the multi-agent case if we are to
represent cooperative and noncooperative interactions, simultaneous
interactions are again required. Thus, we must have a theory of
simultaneous actions for single and multiple agents.

\subsubsection{The Ultrastructure of Time Points}

My solution is to have the agents act in  a virtual sequence, not
in real, macro time but in micro-time. By {\em macro-time} I mean
normal linear time which we will, for simplicity, consider to be
discrete.  {\em Micro-time} is the virtual time that makes up the
fine structure of a time point.  Note, it is not a time in between
two macro time points.  That would lead to problems with dense and
real time since there is no next time point if we have dense or
real time.  Furthermore, micro time is similar to, but not
identical to, an interleaved process used to simulate simultaneous,
parallel actions on a single processor computer.  For that is just
turn taking in small time steps.	 

Instead, we consider the time point itself to have structure.  Its
structure is a tree of micro-actions taken by agents.  We will call
it the {\em micro-time tree} associated with a time $t$.  Any path
in the tree leads to a new state at when the time is over.  With
discrete time this is easier to see.  At any time point $t_1$ an
agent has a finite number of simultaneous {\em micro-choice points}
available.  Each micro-choice point gives the agent a set of
possible micro-actions.

\subsubsection{Micro Actions in Micro Time}

A {\em micro-action} is a primitive act that the agent can do.  At
each choice point the agent must choose one micro-action.  At each
choice point there may also be a {\em do-nothing} or {\em null
micro-action}. The result is a complete subpath in the micro-time
tree which we will call the agent's {\em complete choice}.   We
might call this a {\em complete personal choice} or {\em complete
single-agent choice} to emphasize its individual nature.   Each
agent makes a complete choice by making a micro-choice at each of
his choice points.  The result is a {\em complete multi-agent 
micro-time path} through the multi-agent micro-time tree.  Each
complete path of micro-choices leads to a world state.

\subsubsection{Micro Events in Macro Time}

Actions occur in a world history $H$.  Each micro-action $a_i$ at
time $t$ is realized as an event in a next possible state $K_{t'}$
where $K$ is backwards identical with $H$ at time $t$.  The world
$K$ corresponds to the choice $a_i$ made by the agent.  In fact,
the actual next state will contain a complete choice event
corresponding to the agents complete choice.  A way to think of this
is that actions are first played out in micro-time but then
actually realized as events in macro-time.  In fact the world itself
is one of the agents that makes a complete choice as well, in the
sense that its possibly indeterministic move is part of the
interactivity that generates the next state in a multi-agent system.
The simultaneous actions literally occur within a time point.  The
effect is on the next possible states.

\subsubsection{Simultaneous Actions}

Micro time allows the formalization of simultaneous actions for a
single agent and for multiple agents. Each of the agents may make
their own personal simultaneous micro choices to construct their
simultaneous action or complete choice.  What is important is that
the micro-time tree relates the initial states to the possible next
states.  In doing so, it specifies the interactions that occur
between simultaneous microactions by way of the effect on the
resulting state.  Let us note, we have said nothing about what the
agent actually knows about his micro-choices.  He may make some
micro choices without being aware of it.  And, his choices may only
be partially determined.  We will discuss this case below.

\section{Intentions and Uncertainty}

In a plan based theory of intentional states, we must confront the
fact that agents are uncertain about their own intentions and
plans, as well as the intentions and plans of other agents and
groups of agents.	We now present a theory of plan uncertainty.
This is a theory of how to represent partial information about the
plans and intentions of single agents and groups of agents. The key
step that I made, when I first developed the theory of intentional
states, was to see the analogy between uncertainty about states of
the world and uncertainty about plans and intentions. Thus, we can
represent uncertainty about plans, strategies and intentions as
{\em sets} of possible strategies $S$.  Such a set corresponds to
partial information about the strategy of an agent. Or, from a
different perspective, it corresponds to a partial plan the agent
may be following. Once we have a representation of partial
information about intentions, we can represent perfect information
about intentions, imperfect information about intentions and total
uncertainty about intentions in a way analogous to state
information. Except, now the possibilities are not histories of the
world, but possible strategies that an agent or group might follow.
This allows the formalization of a semantics of ability and
intention where the agents have partial knowledge of intentions.
Furthermore, it turns out that there interesting conceptual and
practical relationships between state information, strategic
information (information about intentions) and
ability~\cite{WernerUniView}. State information creates possible
strategies for an agent. It, thereby, also creates abilities
and enables possible intentions an agent may achieve.

Just as state information reduces the agent's state uncertainty, so
strategic information reduces the agent's plan uncertainty.  An
agent's plan state gives information and describes the uncertainty
about the agent's plans.

\subsection{Strategic Information}

In order to formally define plan uncertainty, we need some formal
preliminaries.  Let {\cal Inf} be the class of all possible
information states. A {\em pure strategy} $\pi$ is a function from
information sets $I$ in {\cal Inf} to the alternatives at $I$.
Thus, $\pi(I) \in Alt(I)$.  {\cal Strat} is the set of all possible
strategies. If $\pi$ is a pure strategy, let  $\pi^{*}$  be those
histories $H$ in $\Omega$ that are compatible with that strategy.
We refer to the class $X^{*}$ generated by the star operator $*$ on
$X$ as the {\em potential} of $X$.   The star operator $*$ denotes
the set of possible histories (past, present, and future) allowed
by the strategy or plan state.  $\pi^{*}$  is then the potential of
a strategy $\pi$  and consists of all those possible histories that
are not excluded by the strategy $\pi$, i.e., \begin{eqnarray}
\pi^* =_{df} \{ H : H \in \pi_{A}(I), \forall  I \in \Xi_{A}(H) \}
\end{eqnarray} where \begin{eqnarray} \Xi_{A}(H) =_{df} \{ I : I
\in \Xi_{A}  and  \exists t \in \Psi, H^{t} \in I \} \end{eqnarray}

Since, action occurs in the context of information  $I$  possessed
by the agent, let $\pi^{*}[I]$ be the set of worlds allowed by the
strategy $\pi$ given the information $I$.   Thus, 
\begin{eqnarray} \pi^{*}[I] =_{df}
\pi^{*}  \cap  I^{*}
\end{eqnarray} 

\subsection{Single Agent Plan Uncertainty}

\subsubsection{Given Perfect State Information}

Recall when $A$ has perfect information his information sets $I$
are unit sets of the form $I(H^t)  = \{H^t\}$. Given perfect
information we need to describe the agent's plan state $S_A$ for
each of $A$'s possible information states $I = \{H^t\}$, or, more
simply, for each state $H^t$.

We describe an agent $A$'s partial plans by a class of strategies
$S_{A}$  that govern that agent's actions.  Any strategy  $\pi$ in
$S_{A}$  is one of the agent's possible strategies that may be
guiding his actions.  Analogous  to information sets, if $\pi$  is
not in $S_{A}$, then it is known  $\pi$  is not guiding $A$'s
actions.  We refer to  $S_{A}(H^{t})$  as the {\em plan state} of
agent $A$ at time $t$ in the world  $H$. More formally:

Let $\Pi_{A}$ be the set of all possible strategies of
agent $A$. For each $H^{t}  \in  V(\Omega)$, let $S_{A}(H^{t})
\subseteq \Pi_{A}$.  $S_{A}(H^{t})$  is the set of $A$'s possible
strategies that are guiding $A$'s actions at time $t$ in the world
$H$. 
\begin{eqnarray} 
S_{A}(H^{t})^{*}  =_{df}  \cup_{\pi \in S_{A}(H^t)} \pi^{*}
\end{eqnarray} 
describes the potential worlds, past, present and future, that are
allowed given strategic information  $S_{A}$ in the world $H$ at
time $t$. $S_{A}(H)$ is the {\em plan history} of the agent $A$ in
the world history $H$. It describes the changing plan states in the
world $H$.

$S_{A}^{B}(H^{t})  \subseteq  \Pi_{B}$  is the strategic
information that agent $A$ has about agent  $B$'s plans in
the world $H$ at the time $t$.  $S_{B}(H^{t})  \subseteq
S_{A}^{B}(H^{t})$  when  $A$'s strategic information is correct.

\begin{eqnarray} 
S_{A}^{B}(H^{t})^{*} =_{df}  \cup_{\pi \in S_A^B(H^t)} \pi^{*}
\end{eqnarray} 
is the set of worlds that are possible given what agent $A$ knows
about $B$'s plans in the world $H$ at time $t$.  Thus,
$S_{A}^{B}(H^{t})$  and  $S_{A}^{B}(H^{t})^{*}$  are two ways of
representing $A$'s strategic information about $B$'s plans $\Pi$
there is perfect information about the state of the world up to time
$t$.

With the above semantic structures we are able to formally
represent the case where agent $A$ may have perfect state
information, but only partial plan, intentional or strategic
information.  

\subsubsection{Given Imperfect State Information}

When an agent $A$ has imperfect state information, we need to
extend the above definitions of knowledge about plan states.  Let
\begin{eqnarray} S_{A}(I) =_{df}  \cup_{H^{t} \in I} S_{A}(H^{t})
\end{eqnarray}   This
represents $A$'s plan state given the partial state information $I$.
Since each world state $H^t \in I$ is indistinguishable for $A$,
each plan state $S_{A}(H^t)$ is identical for each state $H^t$ in
$I$. Hence, $S_{A}(I_A) = S_A(H^t)$ for $H^t$ in $I_A$.

As for $A$'s strategic information about another agent's plan
states, $A$ may be able to distinguish $B$'s plan states over
different world states $I^t$ in $I$. Let  
\begin{eqnarray} 
S_{A}^{B}(I) =_{df}
\cup_{H^{t} \in I} S_{A}^{B}(H^{t})
\end{eqnarray}   
Then
$S_{A}^{B}(I_{A}(H^{t}))$ represents $A$'s strategic information
about  $B$'s  plans relative to  $A$'s state information
$I_{A}(H^{t})$.

\begin{eqnarray} 
S_{A}^{B}(I_{A}(H^{t}))^{*}  =_{df}  \cup_{\pi \in S_A^B(I(H^t))}
\pi^{*}
\end{eqnarray}    
describes the potential worlds (past and future) that
are possible given agent $A$'s strategic information about $B$'s
plans given $A$ has imperfect state information about the world at
time $t$.

We can now represent agents having imperfect state information and
imperfect intentional, strategic information.  Next, we look at the
multi-agent cases.

\subsection{Multi-Agent Plan Uncertainty}

\subsubsection{Given Perfect State Information}

In order for an agent to determine what he can do, the agent must 
be able to represent the plans not just of other individual
agents, but of whole groups of agents.  Let  $Ag$  be a group of
agents,  $Ag  =  \{1,  \ldots  , n\}$.  Then, 
\begin{eqnarray} 
S_{A}^{Ag}(H^{t}) & =_{df} & \{S_{A}^{i}(H^{t})\}_{i \in Ag} \nonumber \\ 
& =\ \   & 
\{<1, S_{A}^{1}(H^{t})>,  \ldots  , <n, S_{A}^{n}(H^{t})>\}
\end{eqnarray}   
This represents the strategic knowledge
that agent $A$ has about the plans of other agents $Ag$ in
the world $H$ at time $t$ given $A$ has perfect state information.
It is the strategic information that the agent $A$ has under the
assumption that  he has perfect information about the state of the
world.  The potential 
\begin{eqnarray} 
S_{A}^{Ag}(H^{t})^{*}  =_{df}
\{S_{A}^{Ag}(H^{t})\}^{*}=_{df} \cap_{i \in Ag}
{S_{A}^{i}(H^{t})}^{*}
\end{eqnarray}   
Note, we take the set intersection of the
potential strategy classes of the individual agents, because those
are the worlds possible given the combined plans of the agents in
$Ag$.

\subsubsection{Given Imperfect State Information}

The {\em plan state} $S_{Ag}(I_{Ag}(H^t))$ of a group of agents 
$Ag$ given imperfect 
information $I_{Ag}(H^t) = \{ I_1(H^t),  \ldots  , I_n(H^t)\}$
is defined as follows:  

\begin{eqnarray} S_{Ag}(I_{Ag}(H^t)) = \{S_1(I_1(H^t)),
\ldots ,  S_n(I_n(H^t))\}
\end{eqnarray}   

Here the potential is defined as follows: 
\begin{eqnarray} 
S_{Ag}(I_{Ag}(H^t))^*
=_{df} \cap_{i \in Ag}S_i(I_i(H^t))^*
\end{eqnarray}   
It is the set of
possibilities allowed by the plan state $S_{Ag}$ of the group given
the state information $I_{Ag}$ of the group.  

To define the strategic information an agent $A$ has about a
group's plans, given his  imperfect state information, we proceed
as follows:

\begin{eqnarray} 
S_{A}^{Ag}(I_{A}(H^{t})) & =_{df} & 
\{S_{A}^{i}(I_{A}(H^{t}))\}_{i \in Ag} \nonumber \\
& = \ \ & 
\{<1, S_{A}^{1}(I_{A}(H^{t}))>,  \ldots  ,
<n,S_{A}^{n}(I_{A}(H^{t}))>\}
\end{eqnarray}  

Here,  $ S_{A}^{Ag}(I_{A}(H^{t}))$ describes agent $A$'s information
about the plans of all other agents $Ag$ given $A$'s state
information $I_A$ in the world $H$ at time $t$.

The potential is defined by:  
\begin{eqnarray} 
S_{A}^{Ag}(I_{A}(H^{t}))^{*}
=_{df} \{S_{A}^{Ag}(I_{A}(H^{t}))\}^{*}=_{df} \cap_{i  \in  Ag}
S_{A}^{i}(I_{A}(H^{t}))^{*}
\end{eqnarray} 
Here, $S_{A}^{Ag}(I_{A}(H^{t}))^{*}$ represents the set of pasts,
presents, and futures that are possible given what agent $A$ knows
about the plans $S_{A}^{Ag}$ of the other agents  $Ag$ relative to
$A$'s state information $I_{A}$ in the world $H$ at time $t$.
Intuitively, it describes the effects of plans of the other agents
given the available information at the time $t$ in the world $H$.
The agent $A$ can then play what-if scenarios about what would
happen if $A$ adopts the plan $\Pi_A$ as part of his plan state.
Then 
\begin{eqnarray} \Pi_A \cap   \{S_{A}^{A}(I_{A}(H^{t}))\}^{*} \cap
S_{A}^{Ag}(I_{A}(H^{t}))^{*}
\end{eqnarray}  
is the set of possible futures of
what would happen if $A$ adopts $\Pi_A$.

\subsection{Plan Consequence Uncertainty}

With any partial or total strategy {\cal P}  we associate, relative
to an agent $A$ with information state $I_A$, a set $\Pi^*$ of
possible worlds that are possible consequences of the strategy.
These are the actually or really possible, consequence worlds of
the strategy given the information available to the agent. An agent
will in general not be able to compute $\Pi^*$.  Instead, the agent
will associate expected consequences (a class of possible events)
with the strategy.  These expectations generate a set of possible
futures $C_{A,I}(\Pi)^{*}$ thought possible by the agent $A$
relative to information $I$.  We say the agent's expectations for
{\cal P} are {\em correct} if and only if $\Pi^* \subseteq C_{A,I}(\Pi)^{*}$.
This states that if an agent does not know the exact consequences of
a strategy, his assessment is correct if it does not contradict the
actual possible consequences of the strategy.

\section{Information and Ability}

When we get up in an unfamiliar hotel room in the middle of the
night, our first reaction is to find the light switch and turn on
the light. Without the light we can do very little.  We might bump
into furniture or we may fall down because of some other obstacle.
When we turn on the light, suddenly all is clear and we can take
control of the situation; we can do great deal more than before. How
is this possible?  You will recall, an increase of information leads
to a decrease in possibilities.

\subsection{Possibilities Generated by Information}

The solution to this problem is that, formally, while the number
of possibilities decreases, the number of strategies available to
an agent increases with an increase of state information.  There
are literally more choices than there are without the
information. Furthermore, the {\em specificity} of the
consequences of those choices increases with more state
information.  A simple example of the game of matching pennies
played once with perfect information and once with imperfect
information makes this obvious (see examples in
\cite{WernerUniView}).

More generally, an increase in state information makes more actions
possible for the agent.  Often a message may only convey state
information such as 'The door is unlocked'.  The receiving agent,
however, has more abilities as the result of the message.  He now
can, for example, 'Open the door' without requiring a key.  Another
example is the function of the senses.  Vision, in humans and
animals, increases the state information about local space-time
tremendously, reducing the state uncertainty, increasing state
information and increasing strategic action possibilities.

In fact, we can view space as an action possibility manifold 
that presents action possibilities to the agent.  As the
philosopher Berkeley observed, vision is the language of God.  It
presents world state information to the agent so that the agent can
act at all.  It was Kant who recognized the human contribution of
the visual apparatus to the construction of the space time manifold
of action possibilities. The action space is constituted by the
state information resulting from visual information as interpreted
by the categories and processes of the mind ( see~\cite{Kant1781}
and later~\cite{BohrWorks,Marr83} ).  Perception in this abstract
sense can be described as an operator on the information state of
the observer.  This state information is used to construct the
action possibility space for the agent.

\subsection{What is the connection between communication and
abilities?}

Communication translates into abilities. We saw there are at least
two basic types of information communicated, state information and
strategic information.  The communication of state information
translates into abilities by making more strategies available to the
agent.  This expands the agents sphere of abilities.

The communication of strategic information can be of two types:
First, strategic information can also create strategies by showing
the agent how to do something. This creates a dynamic, accessible
strategy.  It may have existed as a logically possible strategy in
$\Pi$, however the agent must actually have that strategy in his
dynamic repertoire $\Delta$.  Second, communication of strategic
information can give information about the intentions of other
agents.  This can be utilized by the agent to coordinate, or
cooperate by adjusting his own intentions to achieve his ends.

\section{Types of Ability}

The use of the word 'can' has many senses and its meaning varies in
different contexts of use.  These kinds of ability turn out to be
related to the social context of state and strategic information
that is available and utilized by the agent in his reasoning about
his own and others abilities.  Abilities are also central in agent
reasoning and in the formation of dependencies and power relations
between agents.  Because of their importance, we will analyze
various senses of the word 'can' and more generally the concept of
ability.

The general theory of ability underlying the logic and semantics of
'can' and 'is able to' that I developed is based on a simple
hypothesis, namely: An agent $A$ can do or achieve $\alpha$ if $A$
has a strategy that leads to $\alpha$ being the case.  Many
variants of this basic idea emerge if one considers different
physical and social contexts in which the abilities of single and
multiple agents are ascertain.  Now in ordinary English we sweep
most of these semantic and pragmatic differences under one
syntactic rug. That is to say, we use the same word 'can' to cover a
variety of different semantic and pragmatic meanings.  For clarity
of formalization and implementation of the notion of ability,
however, it is best to be aware of the differences.  Thus, we
artificially construct different types of 'can' operators by giving
them labels to make it easier to recall and distinguish their
meanings.  Yet in ordinary English usage all of these forms are
just referred to by 'can' or 'is able'.

\subsection{Subjective and Objective Ability}

I use the term {\em objective} to mean that which is judged from a
perspective external to the agent.  This may be the perspective of
another agent with information different from that available to
the agent or it may be the meta-perspective of an all knowing
being. This is contrasted with the term {\em subjective} which is
used to refer to the subjective, intrinsic perspective of the agent
himself as limited by his local information whether this be state
information, strategic information or evaluative information.

Throughout this essay we distinguish between the objective
strategies $\Pi_A$ available to an agent $A$ and the subjective
strategies $\Delta_A$ that are actually available given the
agent's actual know-how and limited capacities. The subjective
strategies are those generable dynamically from a 'library' of
substrategies. For example, in a game of chess it may be true that
Boris has an objective strategy $\pi_{Boris}$ in $\Pi_{Boris}$ to
win the game against Bobby, but since Boris is just an average chess
player he does not actually have this strategy available in
$\Delta_{Boris}$. Bobby,  however, is a grand master and knows the
strategy $\pi_{Boris}$ that would allow Boris to win against him.

For any strategy $\pi_A$ that an agent $A$ has $\pi_A$ generates a
set of possibilities that are consistent with that strategy.  We
denote the set of possibilities allowed by the strategy $\pi_A$ by
$\pi_A^*$.  Any world history $H$ in $\pi_A^*$ is a possible
outcome of the strategy.

Let $\alpha$ be any sentence of a language that talks about events
and actions in our multi-agent space-time world.  $\alpha$ may be a
tensed modal formula or a temporally indexed formula. Given any
strategy $\pi_A$ of some agent $A$, we say that strategy is
{\em a strategy for} $\alpha$ if for all possible outcomes $K$ in
$\pi_A^*$, $\alpha$ is realized in $K$. Note, a partial strategy
for $\alpha$ is just a more general version of a strategy for
$\alpha$ that many particular instance strategies may satisfy. Thus,
if $\pi$ is a partial strategy for $\alpha$ then any more specific
instance of $\pi$ will also be a strategy for $\alpha$.

Given these preliminaries about strategies we now define the
objective and subjective abilities of an agent in terms of
available strategies.

\subsection{Objective Ability or Logical Can}

When an agent objectively or logically can do something, then
judged objectively, from an external perspective, there is a
strategy in the logical space of possible strategies for some goal
$\alpha$. The strategy is not necessarily in the space of realizable
dynamic plans $\Delta$. 
 
An agent $A$ {\em objectively can} $\alpha$ at time $t$ in the
world $H$ if $A$ has a strategy $\pi_A$ in $\Pi_A$ and for all
future possibilities $K$ from the time $t$, $\alpha$ holds in $K$.
In effect, this says no matter what the other agents do, a strategy
for $\alpha$ insures that $\alpha$ will hold in any case.

\begin{eqnarray}  \Phi(A \ocan \alpha, H, t) = T \mbox{ iff }
\exists \pi \in \Pi_A \mbox{ such that }  \forall K \in \pi^* ,
\Phi(\alpha, K, t) = T   \end{eqnarray} 

\subsection{Subjective Ability}

The agent has a strategy for $\alpha$ and that strategy is part of
the space of realizable dynamic plans $\Delta$ of the agent.  For
example, Joan is grand master in chess and can see that she can win.
She has a realizable strategy for $\alpha$.	 There is the
subjective component here in that the agent must realize she is
able to do the action based on her available strategies.  

An agent $A$ {\em subjectively can} $\alpha$ at time $t$ in the
world $H$ if $A$ has a strategy $\pi_A$ in $\Delta_A$ and for all
future possibilities $K$ from the time $t$, $\alpha$ holds in $K$.
Like the case of objective ability, no matter what the other agents
do, a subjective strategy for $\alpha$ insures that $\alpha$ will
hold in any case.

\begin{eqnarray}  \Phi(A \scan \alpha, H, t) =
T \mbox{ iff } \exists \pi \in \Delta_A \mbox{ such that
}\nonumber\\ \forall K \in \pi^* , \Phi( \alpha, K, t) = T 
\end{eqnarray} 

Note, that there is an added complexity here, for the agent may
have a dynamic plan for some event, but not realize it.  We	could 
distinguish between what the agent can dynamically do, from what
the agent can dynamically know that he can do.  This distinction is
conflated in the notion of the possible dynamic plans $\Delta$.
Thus, we could distinguish the {\em logically possible abilities}
of the agent from the {\em real abilities} of the agent and this in
turn from the {\em subjective abilities} the agent is aware of that
he has them.  The two types of objective abilities, the logically
possible and the real abilities, is in part due to the fact that
the agent must actually have a representation of the strategy to
able to use it.  Thus, comes in the importance of experience and
training in building the repertoire or basis for dynamically
generating real strategies.

\subsection{Coordinative Ability or \cocan}

In the objective and subjective forms of \can,  if $A \can\ \alpha$
holds in the world $H$ at time $t$, then $\alpha$ will eventually
hold no matter what the other agents do.  A softer version of \can
is not so absolute and depends on the agent's knowledge or
expectations of the intentions of other agents.  The agent utilizes
this knowledge of the intentions of other agents to increase his
ability.

In other words, the agent has a strategy that is dependent for its
success on the intentional plan structure of the social space.  The
strategy by itself with different social information may not
succeed in achieving the goal.  For example,  Henry loves Sue.  He
knows she will be at the chess match at 10 AM tomorrow.  So, Henry
{\sf co-can} meet Sue, because of his knowledge of her intentions.
Sue does not like Henry and would not show up if she knew of
Henry's intentions. But, since Sue does not have this knowledge of
Henry's intentions, Sue is not {\sf co-able} to avoid this.  She
{\sf co-could} avoid Henry if she knew of his intentions.

Indeed, a lot of human effort is spent in trying to determine the
intentions of other persons, groups and nations without them knowing
that their intentions are partially known.  For once those
intentions are known that strategic knowledge gives a form of power
to the knowing agent. This form of power is captured by our
formalization of {\sf co-can}.

\begin{eqnarray}  \Phi(A \cocan \alpha, H, t) =
T \mbox{ iff } \exists \pi \in \Delta_A \mbox{ such that
}\nonumber\\ \forall K \in \pi^* \cap [S_A^{\mbox{Group}}]^*, \Phi(
\alpha, K, t) = T \end{eqnarray}

Here, $\Delta_A$ consists of the possible dynamic plans that are
dynamically constructible from his repertoire of basic plans.
\cocan as well as the forms of \can below have both a subjective
and objective form. For the sake of brevity we present only the
subjective form.

\subsection{Cooperative Ability or Coop-Can}

A group coop-can achieve some goal $\alpha$ if each member of the
group has a strategy such that the combined (possibly coordinated)
space-time effect of the strategies results in the achievement of
$\alpha$. For example,  Joan likes Fred and Fred likes Joan and so
they arrange to meet at the chess tournament at 12 P.M. tomorrow.
Each independently can (in the single agent sense of \can
above) be at the tournament at 12, and the combined effect is the
joint space-time event of meeting at 12.

More generally, mutual communicative interaction by linguistic
communication, gesture and observation of each others actions, can
be used to form the intentional states $S_A$ and $S_B$ of two (or
more) agents $A$ and $B$. The superposition of the resulting
intentional states of a group of cooperating agents can achieve
events that are not achievable by any subgroup, including individual
agents.  These events can be of any form, including helping and
hindering events. Sometimes an event meant to hinder can have the
effect of helping and vice versa.  These are situations repeated
over and over again in novels and films.

\begin{eqnarray} \Phi(\mbox{ Group}_1 \coopcan \alpha, H, t) =  T
\mbox{ iff } \forall A \in \mbox{ Group}_1 \ \exists \pi_A \in
\Delta_A \mbox{ such that }\nonumber\\ 
\forall K \in \cap_{A \in
Group} \pi_A^*, \Phi( \alpha, K, t) =	T \nonumber 
\end{eqnarray}

This states that group {\em Group}$_1$ \coopcan the event $\alpha$
if they have strategies whose superimposed interaction results in
$\alpha$ being achieved.  This means that $\alpha$ is inevitable if
the agents set up their strategies using communication of some form.
Below we will refer to \coopcan$_{Group}$ by  \can$_{Group}$ where
single agent can is just a special case of group can.

\subsection{Coordinative Cooperative Ability or Co-Coop-Can}

A group {\em Group}$_1$ can cooperatively achieve some event
$\alpha$ using their knowledge of the intentions of other groups of
agents {\em Group}$_2$. In fact, each agent $A$ of group {\em
Group}$_1$ uses his individual strategic knowledge of the intentions
of group {\em Group}$_2$.  In this case we have a form of
\cocoopcan.  The agents can interact to form intentions that
superimposed will interact with the co-intentions of a second group
{\em Group}$_2$ to achieve their ends.

\begin{eqnarray}  \Phi(\mbox{ Group}_1 \cocoopcan
\alpha, H, t) = T \mbox{ iff } \forall A \in \mbox{ Group}_1\
\exists \pi_A \in \Delta_A \mbox{ such that }\nonumber \\  
\forall K  \in \cap_{A \in \mbox{Group}_1} \pi_A^* \cap
[S_A^{\mbox{Group}_2}]^*, \Phi( \alpha, K, t) = T 	  \nonumber
\end{eqnarray} 

Note, \can in all of its above forms only says that the
appropriate subjective and objective strategies exist. It does not
say or imply that such strategies will be adopted and, hence, does
not imply that the corresponding strategic intentional states will
be formed.  Indeed, in the case of \coopcan there may be a heavy
coordination cost associated with the formation of such
interlocking strategic intentional states.

\section{A Theory of Utilitarian Ability}

The above forms of ability of single and multiple agents do not
consider the cost or utility of performing or achieving some event
described by $\alpha$.  We now formalize forms of ability that make
the utility an explicit part of the operator.  We restrict our
discussion to the subjective and objective forms of \can.  The
\cocan and \coopcan forms also exist.

Let us denote the utility that $\alpha$ has for an agent $A$ in the
world $H$ at time $t$ by $U_A(\alpha, H, t)$.  Thus, the utility of
some event described by $\alpha$ is relative to the agent $A$, the
history $H$, and the time $t$. If the utility is negative,
$U_A(\alpha, H ,t) < 0$, then we call the utility the
{\em cost} or {\em negative utility} of $\alpha$. If
the utility is positive, it is called the {\em value}
or {\em positive utility} of $\alpha$ for the agent
$A$. If the utility is equal to $0$ we say that
$\alpha$ has {\em no utility} for $A$ or that the utility is {\em
neutral} or that there is {\em no cost} or {\em no value} for $A$.

\subsection{Pessimistic but Safe Utilitarian Ability}

For an strategy $\pi$, we say $\pi$ {\em forces} $\alpha$, in
symbols $\pi \forces \alpha$ if and only if for all $H \in \pi^*$,
$\alpha$ holds in $H$. Let $\Psi_t = \{ t': t' \in \Psi \mbox{ and
} t' > t\}$.  Then we can define utilitarian ability as follows:
$\Forces$

\begin{eqnarray}  A \can_t^{u}\ \alpha(t') \mbox{ in } H \mbox{ iff
} \exists \mbox{ strategy } \pi \in \Pi_A \mbox{ such that } \pi
\forces \alpha \mbox{ and the utility }\nonumber\\ u = \mbox{\bf
min }_{H \in \pi^*,\, t' \in \Psi_t}[U_A(\alpha(t'), H)]
\end{eqnarray} 

Note, this gives the minimal positive utility or value
or the maximal possible negative utility or cost of attaining
$\alpha$ given the strategy $\pi_A$ is followed.  Thus, this sense
of utility relativized ability is a {\em pessimistic} but {\em
safe} minimal damage assessment.  If the utility $u$ is positive
then it is the minimum that the agent can gain if he follows the
strategy $\pi_A$.  If the utility is negative, representing a cost,
then it is the most the agent can loose given that he follows the
strategy.  

\subsection{Optimistic but Unsafe Ability}

An {\em optimistic} but unsafe assessment of ability is obtained if
we substitute {\em max} for {\em min} in our semantic definition.

\begin{eqnarray}  A \can_t^{u}\ \alpha(t') \mbox{ in } H \mbox{ iff }  \exists
\mbox{ strategy } \pi \in \Pi_A \mbox{ such that } \pi \forces
\alpha \mbox{ and  the utility }\nonumber\\  
u = \mbox{\bf max }_{H \in \pi^*,\, t' \in \Psi_t}[U_A(\alpha(t'),
H)] 
\end{eqnarray} 

If the optimistic utility $u$ is positive it represents the
greatest possible benefit that may result from the strategy $\pi$.
If the optimistic utility $u$ is negative, it is the least costly
result possible under the strategy $\pi$.

\subsection{Probabilistic Ability}

Often agents act with the outcome of the action being only probable
and not certain.  This is actually the most prevalent case for
everyday human action.  It is, therefore, interesting to formalize
this more realistic sense of ability.  In probabilistic ability an
agent can do $\alpha$ with probability $p$ if there is a strategy
$\pi_A$ that insures $\alpha$ will be the case with probability $p$.
Let $\alpha(\pi^*_{A}) = \{K : K \in \pi^*_A \mbox{ and } \alpha
\mbox{ holds in } K\}$. Intuitively,  $\alpha(\pi^*_A)$ is the set
of futures allowed by the strategy where $\alpha$ is true.
\begin{eqnarray} A \can_{t}^{p}\ \alpha \mbox{ in } H \mbox{ iff } \exists \mbox{
strategy } \pi \in \Pi_A \mbox{ such that } \sum_{K \in \pi_A^*
\cup\, \alpha(pi^*_A)} p(K) = p \end{eqnarray}  Here, $p(K)$ is the
probability that the possible future $K$ will be actually realized.
		   
\subsection{Probabilistic Expected Utilitarian Ability}

Assume we have a probability distribution $p$ over the choices of
the agents where $\sum_{K \in \pi_A^*} p(K) = 1$. Furthermore, we
say a strategy $\pi_A$ {\em forces} an event $\alpha$ {\em with
probability} $p$, in symbols, $\pi \pforces \alpha$, if an only if
$\sum_{K \in \pi_A^*} p(K) = p$.
Then we can define
the expected utility of $\alpha$ in the world $H$ at time $t$ as
follows:

\begin{eqnarray}  A \can_t^{xu}\ \alpha(t') \mbox{ in } H \mbox{ iff }  \exists
\mbox{ strategy } \pi \in \Pi_A \mbox{ such that } \pi \pforces
\alpha \mbox{ and the expected utility }\nonumber\\  xu = \sum_{K
\in \pi_A^*} p(K) U(\alpha(t'), K) \end{eqnarray}

This says that the agent $A$ can do $\alpha$ and can expect to have
a payoff or cost depending on whether the expected utility $xu$ of
$\alpha$ is positive or negative, respectively.  The strategy
$\pi_A$ generates a set of possible futures which we denote by
$\pi_A^*$.  This set of futures has probabilities associated with
each possible course of action $H$.  These probabilities are then
used to calculate the average or expected utility of the
action-event described by $\alpha$.

The expected utility can be a positive benefit or a negative cost
or a neutral value of zero.  Thus, if $A \can_t^{xu}\ \alpha$
and $xu$ is negative then agent $A$ can do $\alpha$ but he expects
it to cost him $xu$.  If $xu$ is positive $A$ expects to gain
benefit $xu$. And, if $xu$ is zero, then $A$ expects no benefit but
also no cost in doing $alpha$.  Note, since we are dealing with both
probabilities and utilities, very unlikely but very costly events
may have more weight than a less costly and more likely event.  The
point is that the utility can be positive in one possible future and
negative in another possible future.  Which future has more
influence on the expected utility depends on the probability
multiplied by the utility of $\alpha$ in that future.

Probabilistic utilitarian ability is less pessimistic than
pessimistic utilitarian ability and less optimistic than optimistic
utilitarian ability. Thus, probabilistic utilitarian ability lies
between pessimistic and optimistic utilitarian ability.  A true and
careful pessimist will not use expected utilities to assess his
abilities.  For no matter how unlikely, a very costly possible
outcome is still possible even if the expected utility is very
good.

Our motivation for defining the above ability operators is because
they are useful.  First, they reflect a more realistic and
pragmatic approach to ability.  Second, they are useful for
understanding dependency and power relations between agents.

\section{Types of Intention}

Throughout, for each operator we distinguish the consequences of a
plan state that are independent of the actions of other agents and
the consequences that depend on the social intentional state of the
social space.  We add the {\sc co-} prefix to the operator if the
social context is part of the computation of its consequences.

\subsection{Intends or Plans}

For a single agent,  $A$ \plans $\alpha$ if $A$'s intentional state
in the world $H$ at time $t$ relative to his information state $I$
leads to the achievement of $\alpha$ no matter what the other
agents do. 

\begin{eqnarray}  A \plans_t\ \alpha(t') \mbox{ in } H \mbox{ iff } \forall K \in S_{A}(I_A(H^t))^*,\ 
\alpha(t') \mbox{ holds in } H \end{eqnarray}

Here, \plans implies \can.  To make this realistic,
background assumptions about the expected circumstances are usually
added.  An example might be Boris in a game of chess plans to win if
his plan state that is guiding him realizes a winning strategy.
Note, this sense of intends and plans is neutral with respect to
beliefs, desires and awareness.  But, it is dependent on the
information and plan state of the agent.

\subsection{Coordinated Intentions}

For single agents, the agent plans to achieve $\alpha$ by using his
information about the intentions of others.  

\begin{eqnarray}  A \coplans_t\ \alpha(t') \mbox{ in } H
\mbox{ iff } \nonumber\\ \forall K \in 
S_{A}(I_{A}(H^{t}))^{*} \cap S_{A}^{Ag}(I_{A}(H^{t}))^{*},
\alpha(t') \mbox{ holds in } H \end{eqnarray}

The agent is in an
intentional state such that in conjunction or in superposition with
the social space in which the agent acts, the intentional state of
the agent plus the intentional structure of the social space results
in the realization of the event described by $\alpha$.  

\subsection{Cooperatively Intending}

A group of agents cooperatively intends to achieve some event
$\alpha$ if they each have the appropriate intentional states such
that when these intentional states are applied or executed, the
event described by $\alpha$ is realized.

\begin{eqnarray}  \mbox{Group} \plans_t \alpha(t') \mbox{ in } H
\mbox{ iff }  \nonumber\\
\forall K \in S_{\mbox{Group}}(I_{\mbox{Group}}(H^{t}))^{*},
\alpha(t') \mbox{ holds in } H \end{eqnarray}

Cooperative intentions are different from coordinate intentions, in
that cooperative intentions require the active cooperation of the
agents involved. Whereas coordinate intentions can be executed by a
single agent who makes use of his knowledge of the social space,
such as, for example, his knowledge of the intentions of others.

\subsection{Third Person Predictions}

This might be called the 'will result in' operator.  For it has to
do with the logical and physical consequences of a plan that the
agent himself may not be aware of, but which another agent may be
aware of because of additional strategic or state information.  It
is meant to be neutral with regard to the intentions of the agent.
For example, when Sue goes to the tournament she does not realize
that one consequence will be that she {\sf will} run into Henry
whom she wants to avoid.  Note, here this is a consequence in the
social context of other intentions, such as Henry's and that of the
tournament organizers. It is a social context about which our agent,
Sue, does not possess perfect intentional information.
\begin{eqnarray} B \mbox{ has the information that } A \mbox{ {\sf
will}}_t\ \alpha(t') \mbox{ in } H \mbox{ iff }
\nonumber\\ \forall K \in S_{B}^{A}(I_{B}^{A}(H^t))^*, \alpha(t')
\mbox{ holds in } H \end{eqnarray}  
This states that from the
perspective of agent $B$, agent $A$ will do $\alpha$ at time $t'$.
Note, the agent $A$ may make assessments about what he himself will
do.  In this case $B = A$. 

\subsection{Probabilistic Intentions}

Assume that we have a probability distribution over the possible
outcomes of a partial strategy $S_A$.   Let $\alpha(S_{A}^*) = \{K
: K \in S_A^* \mbox{ and } \alpha \mbox{ holds in } K\}$.
Intuitively, $\alpha(S_A^*)$ is the set of futures allowed by the
strategy where $\alpha$ is true.  A partial strategy $S_A$ {\em
forces} an event $\alpha$ {\em with probability} $p$, in symbols,
$S_A \pforces \alpha$, if an only if $\sum_{K \in S_A^* \cup
\alpha(S_{A}^*)} p(K) = p$.  In other words, a partial strategy
forces $\alpha$ with probability $p$ if, under the constraints of
the strategy, the sum of the probabilities where $\alpha$ holds is
equal to $p$.  Given this definition, we can then define the
probabilistic plan state of an agent as follows:

\begin{eqnarray} A \mbox{ {\sf
plans}}_t^p\ \alpha(t') \mbox{ in } H \mbox{ iff
} S_A \pforces \alpha(t')
\end{eqnarray} 

\subsection{Utilitarian Intentions}

People often break their best intentions.  Often, when the path
becomes too difficult, or another alternative is too tempting, some
people stray from promises, commitments and intentions. 

\begin{eqnarray} A \mbox{ {\sf
plans}}_t^u\ \alpha(t') \mbox{ in } H \mbox{ iff }
\forall K \in S_A \mbox{ if } U(\alpha, K) \geq u \mbox{ then	}
\alpha(t') \mbox{ holds in } K  \end{eqnarray}

Thus, an agent with a utilitarian intention is committed to $\alpha$
only to the degree that the utility does not decrease below the
value $u$.  This means an unexpected cost or an enticement can lead
the agent estray from his intention to do $\alpha$.

\section{The Entropy of Plans and Organizations}

Fascinating is the thought of being able to give a measure of the
uncertainty of control in a multi-agent system.  Put differently,
can we come up with a measure of the control information in a
multi-agent organization?  Can we come up with measures of the
entropy of an organization?  We now take some preliminary steps
that take us in the direction of measuring the entropy of
multi-agent systems.

Beyond defining state uncertainty, we have also been able to
define control uncertainty formally and explicitly. Historically,
a formal account of state uncertainty made a formal definition of
the {\em entropy} of a system possible. Thus, given a set of
states $\sigma \in I$ and a probability distribution $p$ over $I$
we can define the entropy of $I$ to be $H(I) = - \sum_{\sigma \in
I} p(\sigma) \log p(\sigma) $. This definition was, in its
essence, first given by Boltzmann~\cite{BoltzmannWorks} in his
statistical foundations of thermodynamics~\cite{Khinchin49} and
later by many others including Shannon~\cite{Shannon48} in his
mathematical theory of communication~\cite{Khinchin57}. Without
probabilities, the entropy is simply the log of the magnitude of
the set $I$.

\subsection{Entropy of Intentional States}

Because we have constructed a formal definition of plan or
control uncertainty, we can now also give an analogous definition
of {\em control entropy}:
 
\begin{eqnarray}  H(S) = - \sum_{\pi \in S} p(\pi) \log p(\pi)
\end{eqnarray}  

where $p(\pi)$ is the probability that the agent is following
the strategy $\pi$ and $S$ is the control information of the
agent\footnote{The strategic information or plan state $S$ is
not to be confused with the "$S$" in the Boltzmann formula for
state entropy:  $S = k \ln P$ where $P$ is the thermodynamic
probability of the system being in a given state and $k$ is
the Boltzmann constant~\cite{BoltzmannWorks}. Furthermore, it
should be obvious from the context that the notation $H()$ for
entropy and the notation $H$ for world history are two very
different things.}. This gives a mathematical measure of the
control uncertainty of an agent. Alternatively, it gives a
measure of the {\em control information content} if that
control uncertainty were removed. To distinguish the new
concept of control entropy from the traditional notion of
entropy, we call the former {\em control entropy} (also, {\em
plan entropy}) and the latter {\em state entropy} since it is
concerned with measuring state uncertainty or, positively, the
state information content if that uncertainty is removed. The
plan state $S$ may be for a single agent or for a whole
multi-agent system.

\subsection{Conditional Control Entropy}

Given the definition of control entropy we can define other
concepts concerning control information that are analogous to
some of the traditional concepts in Shannon's communication
theory. Thus, given two agents $A$ and $B$\footnote{It should be
clear from the above that these may be groups as well.}, with plan
states $S_A$ and $S_B$, the {\em conditional control entropy} is
defined as: 
\begin{eqnarray} H_{S_A}(S_B) = \sum_{\pi \in S_A}
p(\pi)H_{\pi}(S_B) 
\end{eqnarray}  
where 
\begin{eqnarray}
H_{\pi}(S_B) = - \sum_{\theta \in S_B} p_{\pi}(\theta) \log
p_{\pi}(\theta) 
\end{eqnarray} 
where $p_{\pi}(\theta)$ is the
conditional probability that $B$ has plan $\theta$ when $A$ has
plan $\pi$. Expressed in terms of control information $H_A(B)$
indicates how much control information is contained on the average
in the partial plan state $S_B$ given it is known that agent $A$
is following plan state $S_A$.

\subsection{Strategic Entropy Due to Multi-Agent Side Effects}

We have already seen above that actions and strategies, in the
context of a social world of other agents, are not fully
determined. We can actually measure the entropy of a strategy due
to multi-agent side effects.

\begin{eqnarray}  H(\pi_A) = - \sum_{K \in \pi^*} p(K) \log p(K) 
\end{eqnarray} 

where $\pi_A$ is a strategy of agent $A$ and $p(K)$ is the
probability of the future world $K$ given the strategy $\pi_A$.
This measures the uncertainty of the strategy given no
information about the strategies of the other agents.  The
relations between state entropy, plan entropy and strategic
entropy will be the subject of another paper.

\section{Conclusion}

We have tried to understand the relationship of an agent to his
social space. How is coordination and cooperation in social
activity possible?  How is the information about the state of the
world related to the agents intentions?  How are the intentions of
agents related?  What role does communication play in this
coordination process?  Such questions led to fundamental
hypotheses and theories about the nature of social agents.  

What is an agent? By an agent I mean an entity that is guided by
some strategy that controls its behavior.  We have seen that
there are many different types of agents possible.  But the core
of all agents reduces to three fundamental components:  The
information state $I$ that gives the agents information about the
world, the strategic state $S$ that gives the agents control
information that guides his actions, and an evaluative state $V$
that gives the agents evaluations and guides his reasoning. A
solipsistic agent need not even have external inputs.  In simple
agents, the strategic state may simply be a reactive strategy that
is directly linked to the agent's information state, with the
information state being determined by immediate perceptions.  The
evaluative component may be nonexistent. As the agents gain more
social abilities, the representational states of the agent become
more and more complex. The agent now has dynamic properties
constraining the information and strategic states including
properties such as nondiminishing information and perfect
information.

We also investigated the abilities of the agent and saw that what
an agent can do is definable in terms of the existence of
strategies in various contexts of state, strategic and social
information. The existence of strategies and, therefore, what an
agent or group can do and intend, depends on the available state
information. Thus, communication of state information was seen to
have a fundamentally important function for individual and social
action. The intentions of an agent and a group of agents was
investigated and the relationship to state information and other's
intentions was shown. The communication of strategic information has
several important functions in the social space of agents:  First,
strategic information can create real abilities from logically
possible abilities through training or by simply making the agent
aware of their existence.  Second,  strategic information creates
new coordinate abilities since the agent can use strategic
information about other agents to plan his actions.  Third,
strategic information creates cooperative abilities, leading to
the possibility of cooperation in multi-agent activity.  This is
because the communication of strategic information is a
fundamentally important function in the formation of coordinated
intentional states between agents, and, thereby, forms the basis of
cooperation.

The formal modeling of state and plan uncertainty is important for a
deeper understanding of the social space of agents including the
processes of communication, cooperation, and coordination. This has
a direct effect on the design and implementation of single agents
and of multi-agent systems. In this chapter, we hope we have gone
further in the direction of providing a conceptual and logical
foundation for work in the area of planning and distributed
artificial intelligence. In addition, we have extended the concept
of entropy to include the entropy of intentional states. The entropy
of intentions may provide a measure of the entropy of organizations
in their dynamic activity. Multi-agent system science and the
formalization of social processes between agents will, I believe,
lead to a revolution in the psychological, social, economic, and
biological sciences.  By providing a new foundation for these
sciences we hope to contribute to the future understanding and creation of social
beings and multi-agent social systems.






\end{document}